\documentclass{article} 
\usepackage[table]{xcolor}
\usepackage[preprint]{colm2025_conference}

\usepackage{microtype}
\usepackage{hyperref}
\usepackage{url}
\usepackage{booktabs}

\usepackage{lineno}

\definecolor{darkblue}{rgb}{0, 0, 0.5}
\hypersetup{colorlinks=true, citecolor=darkblue, linkcolor=darkblue, urlcolor=darkblue}

\title{Doc2Agent: Scalable Generation of Tool-Using Agents from API Documentation}

\author{Xinyi Ni, Haonan Jian, Qiuyang Wang, Vedanshi Chetan Shah \& Pengyu Hong \\
Michtom School of Computer Science\\
Brandeis University\\
Waltham, MA 02453, USA \\
\texttt{\{xinyini, hongpeng\}@brandeis.edu} \\
}

\usepackage{wrapfig}
\usepackage{graphicx}
\usepackage{makecell}
\usepackage{amsmath}
\usepackage{amssymb}
\usepackage{booktabs}
\usepackage{subfigure}
\usepackage{multirow}
\usepackage{listings}
\usepackage{tcolorbox}
\usepackage{xcolor}
\usepackage{stfloats} 
\usepackage{pgf}

\newcommand{\colorcell}[1]{%
  \pgfmathsetmacro{\percent}{#1}
  \pgfmathsetmacro{\clampedpercent}{max(0, min(100, \percent))} 
  \pgfmathsetmacro{\greenval}{round(100 * \clampedpercent / 100)} 
  \pgfmathtruncatemacro{\greenvalint}{\greenval} 
  \edef\colorcmd{green!\greenvalint!white}%
  \expandafter\cellcolor\expandafter{\colorcmd}%
  $\uparrow$#1\%%
}

\definecolor{jsonbg}{RGB}{248,248,248}
\definecolor{jsonstring}{RGB}{42,161,152}
\definecolor{jsonnumber}{RGB}{211,54,130}
\definecolor{jsonkeyword}{RGB}{108,113,196}
\definecolor{jsoncomment}{RGB}{147,161,161}
\definecolor{jsondelim}{RGB}{38,139,210}

\lstdefinelanguage{JSON}{
  keywords={true,false,null},
  keywordstyle=\color{jsonkeyword},
  sensitive=false,
  comment=[l]{//},
  morecomment=[s]{/*}{*/},
  commentstyle=\color{jsoncomment},
  string=[b]",
  stringstyle=\color{jsonstring},
}

\lstdefinestyle{json}{
  language=JSON,
  backgroundcolor=\color{jsonbg},
  basicstyle=\ttfamily\small,
  breaklines=true,
  breakatwhitespace=true,
  captionpos=b,
  keepspaces=true,
  numbers=left,
  numbersep=5pt,
  numberstyle=\tiny\color{jsoncomment},
  showspaces=false,
  showstringspaces=false,
  showtabs=false,
  tabsize=2,
  literate=
    *{:}{{{\color{jsondelim}{:}}}}{1}
    {,}{{{\color{jsondelim}{,}}}}{1}
    {[}{{{\color{jsondelim}{[}}}}{1}
    {]}{{{\color{jsondelim}{]}}}}{1}
    {\{}{{{\color{jsondelim}{\{}}}}{1}
    {\}}{{{\color{jsondelim}{\}}}}}{1}
    {0}{{{\color{jsonnumber}{0}}}}{1}
    {1}{{{\color{jsonnumber}{1}}}}{1}
    {2}{{{\color{jsonnumber}{2}}}}{1}
    {3}{{{\color{jsonnumber}{3}}}}{1}
    {4}{{{\color{jsonnumber}{4}}}}{1}
    {5}{{{\color{jsonnumber}{5}}}}{1}
    {6}{{{\color{jsonnumber}{6}}}}{1}
    {7}{{{\color{jsonnumber}{7}}}}{1}
    {8}{{{\color{jsonnumber}{8}}}}{1}
    {9}{{{\color{jsonnumber}{9}}}}{1}
}

\usepackage{color}
\definecolor{darkred}{rgb}{0.6,0.0,0.0}
\definecolor{darkgreen}{rgb}{0,0.50,0}
\definecolor{lightblue}{rgb}{0.0,0.42,0.91}
\definecolor{orange}{rgb}{0.99,0.48,0.13}
\definecolor{grass}{rgb}{0.18,0.80,0.18}
\definecolor{pink}{rgb}{0.97,0.15,0.45}

\lstdefinestyle{colored}{ %
  basicstyle=\ttfamily,
  backgroundcolor=\color{white},
  commentstyle=\color{green}\itshape,
  keywordstyle=\color{blue}\bfseries\itshape,
  stringstyle=\color{red},
}
\lstdefinelanguage{PythonPlus}[]{Python}{
  morekeywords=[1]{,as,assert,nonlocal,with,yield,self,True,False,None,} 
  morekeywords=[2]{,__init__,__add__,__mul__,__div__,__sub__,__call__,__getitem__,__setitem__,__eq__,__ne__,__nonzero__,__rmul__,__radd__,__repr__,__str__,__get__,__truediv__,__pow__,__name__,__future__,__all__,}, 
  morekeywords=[3]{,object,type,isinstance,copy,deepcopy,zip,enumerate,reversed,list,set,len,dict,tuple,range,xrange,append,execfile,real,imag,reduce,str,repr,}, 
  morekeywords=[4]{,Exception,NameError,IndexError,SyntaxError,TypeError,ValueError,OverflowError,ZeroDivisionError,}, 
  morekeywords=[5]{,ode,fsolve,sqrt,exp,sin,cos,arctan,arctan2,arccos,pi, array,norm,solve,dot,arange,isscalar,max,sum,flatten,shape,reshape,find,any,all,abs,plot,linspace,legend,quad,polyval,polyfit,hstack,concatenate,vstack,column_stack,empty,zeros,ones,rand,vander,grid,pcolor,eig,eigs,eigvals,svd,qr,tan,det,logspace,roll,min,mean,cumsum,cumprod,diff,vectorize,lstsq,cla,eye,xlabel,ylabel,squeeze,}, 
}
\lstdefinelanguage{PyBrIM}[]{PythonPlus}{
  emph={d,E,a,Fc28,Fy,Fu,D,des,supplier,Material,Rectangle,PyElmt},
}
\lstdefinestyle{colorEX}{
  basicstyle=\ttfamily,
  backgroundcolor=\color{white},
  commentstyle=\color{darkgreen}\slshape,
  keywordstyle=\color{blue}\bfseries\itshape,
  keywordstyle=[2]\color{blue}\bfseries,
  keywordstyle=[3]\color{grass},
  keywordstyle=[4]\color{red},
  keywordstyle=[5]\color{orange},
  stringstyle=\color{darkred},
  emphstyle=\color{pink}\underbar,
}
\begin{document}

\ifcolmsubmission
\linenumbers
\fi

\maketitle

\begin{abstract}
REST APIs play important roles in enriching the action space of web agents, yet most API-based agents rely on curated and uniform toolsets that do not reflect the complexity of real-world APIs. Building tool-using agents for arbitrary domains remains a major challenge, as it requires reading unstructured API documentation, testing APIs and inferring correct parameters. We propose Doc2Agent, a scalable pipeline to build agents that can call Python-based tools generated from API documentation. Doc2Agent generates executable tools from API documentations and iteratively refines them using a code agent. We evaluate our approach on real-world APIs, WebArena APIs, and research APIs, producing validated tools. We achieved a 55\% relative performance improvement with 90\% lower cost compared to direct API calling on WebArena benchmark. A domain-specific agent built for glycomaterial science further demonstrates the pipeline’s adaptability to complex, knowledge-rich tasks. Doc2Agent offers a generalizable solution for building tool agents from unstructured API documentation at scale.
\end{abstract}

\section{Introduction}
Tool agents~(\citet{ferrag2025llmreasoningautonomousai}, \citet{yehudai2025surveyevaluationllmbasedagents}, \citet{Wang_2024}, \citet{qu2025tool}), built on LLMs, aim at interacting autonomously with existing software or web services. One important tool source to enrich the capability of tool agents is from existing \textbf{REST API services}~\citep{barry2003web}. REST APIs are a widely adopted standard for communication between clients and servers, which expose programmic access to web-based services through structured endpoints. API agents~(\citet{qin2023toolllmfacilitatinglargelanguage}, \citet{du2024anytoolselfreflectivehierarchicalagents}, \citet{song2025browsingapibasedwebagents}) use REST APIs by sending HTTP requests with appropriate parameters to access, manipulate, or retrieve structured data from web services. The method to connect REST APIs to these agents varies. For example, ToolLlama~\citep{qin2023toolllmfacilitatinglargelanguage} created a scraper to fetch 16,000 APIs from \href{https://rapidapi.com/hub}{RapidAPI} which provides uniformed API specifications. API-based agent~\citep{song2025browsingapibasedwebagents}, added API documentation as part of input prompt to the agent. On the other hand, API-calling benchmarks such as WebArena~\citep{zhou2024webarenarealisticwebenvironment}, Appworld~\citep{trivedi2024appworldcontrollableworldapps}, ComplexFuncBench~\citep{zhong2025complexfuncbenchexploringmultistepconstrained} provides ready-to-use APIs in an uniform format, usually a predefined JSON schema. While existing benchmarks and agent frameworks typically offer APIs in clean, uniform formats to facilitate convenient function calling, this setup overlooks the real-world challenges posed by inconsistent and low-quality API documentation. The assumption that APIs can be seamlessly used by AI agents does not hold in practice. Beyond well-maintained commercial platforms such as \href{https://rapidapi.com/hub}{RapidAPI} and \href{https://www.postman.com/explore}{Postman}, many APIs lack comprehensive documentation and often do not follow standardized schemas. Even when schemas are available, they may be incomplete or omit critical information, making it difficult to automate tool generation reliably. Furthermore, API services and their documentation are frequently outdated, requiring extensive validation and refinement to ensure usability. These challenges reveal a significant gap between tool agents and real-world API services.

\begin{figure}[htbp] 
    \centering
    \includegraphics[width=0.45\textwidth]{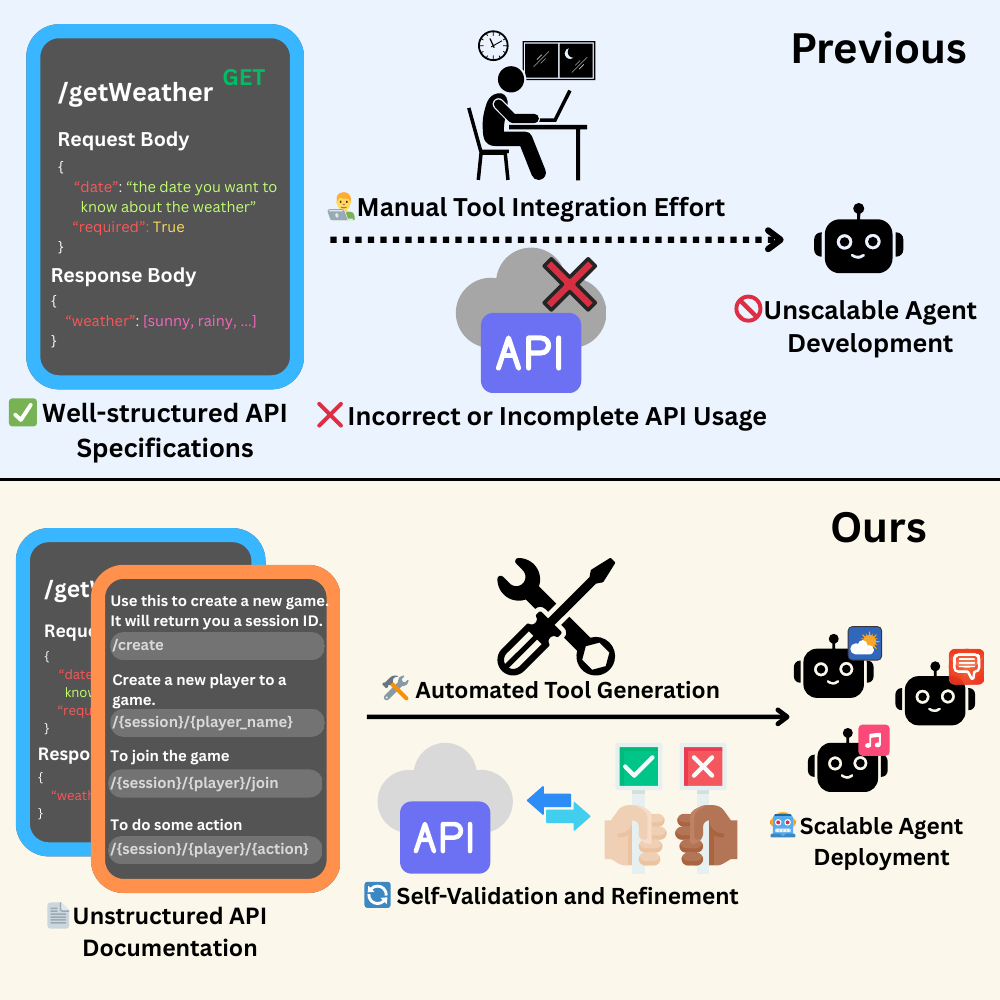}
    \caption{\textbf{Bridging the Gap Between Real-World APIs and AI Agents} (Top) Conventional agent development relies on high-quality APIs and manual integration, leading to scalability issues. (Bottom) Our approach automates tool generation from natural language API documentation, enabling self-validation, refinement, and seamless deployment of AI agents.}
    \label{fig:motivation}
\end{figure}
We argue that it is essential to \textbf{\textit{develop an automated pipeline for scalable agent generation with AI-ready tools from any domain-specific REST APIs}} (Figure \ref{fig:motivation}). Such a pipeline enables rapid tool creation directly from natural language API documentation, enhancing development efficiency for AI practitioners. For end users, API agents provide an intuitive natural language interface to interact with APIs, lowering the barrier to entry. For the AI agent community, this automation facilitates seamless integration of diverse APIs, expanding agent capabilities and supporting the creation of tool-accessible benchmarks. In this work, we focus on constructing Python-based tools that can be seamlessly generated and executed by an action parser~\citep{wang2024executablecodeactionselicit}. These tools are natively compatible with popular agentic frameworks such as \href{https://www.langchain.com/langgraph}{LangGraph}, \href{https://www.llamaindex.ai/}{LlamaIndex}, and \href{https://microsoft.github.io/autogen/stable//index.html}{AutoGen}. Compared to approaches that rely on direct API requests~\citep{song2025browsingapibasedwebagents}, encapsulating functionality as Python functions improves token efficiency and allows code agents to concentrate on task logic and parameter usage, rather than low-level API construction. Another important application of agent-generation pipeline lies in the development of scientific agents~\citep{Wang_2024}. Agents are increasingly demonstrating their versatility across research domains, including science~(\cite{baek2024researchagentiterativeresearchidea, chen2025scienceagentbenchrigorousassessmentlanguage}), healthcare~(\cite{abbasian2024conversationalhealthagentspersonalized}), and finance~(\cite{li2023tradinggptmultiagentlayeredmemory}). These agents are designed to tackle domain-specific challenges—for example, ChemCrow~\citep{bran2023chemcrowaugmentinglargelanguagemodels}, a chemistry-focused research agent, uses expert-curated tools to perform tasks such as data retrieval, analysis, and prediction. However, its tools were manually implemented by researchers. Many research services are already accessible via REST APIs, including dataset repositories~(\cite{sayers2021database}, \cite{rose2021rcsb}) and scientific applications~(\cite{dorst2023nmr}, \cite{Glycam}). Our pipeline has the potential to automate tool generation for such APIs, accelerating the deployment of domain-specific research agents.

Research APIs are often less actively maintained than commercial APIs due to limited developer resources, resulting in lower documentation quality and inconsistent schema design. Furthermore, research agents typically need to query across multiple datasets, each using distinct entries, representations, and semantic conventions. These variations pose significant challenges for using correct input parameter, especially in the absence of domain-specific prior knowledge.

To address the limitations in tool agent development, we propose an open-source pipeline, \textbf{Doc2Agent}\footnote{Code available at \href{https://github.com/coolkillercat/Doc2Agent}{\url{https://github.com/coolkillercat/Doc2Agent}}}(Figure \ref{fig:pipeline}), which \textbf{(1)} Autonomous generation of Python-based tools from any REST API documentation written in natural language, \textbf{(2)} Automatic evaluation of tool functionality to verify operability and alignment with API descriptions, \textbf{(3)} Iterative tool refinement to improve reliability and usability, infer parameter values without requiring prior domain knowledge. \textbf{(4)} Deploy toolkits to MCP server for seamless deployment into any agentic framework.

To demonstrate the effectiveness of our approach, we collected 167 real-world API documentation pages comprising 744 publicly accessible endpoints, 174 API documentations for WebArena benchmark and 16 websites for glycoscience research~\citep{f91d0fc0f6f944c0a108d450c6aa7490}. Differ from prior benchmark datasets, these API docs have diverse and inconsistent documentation formats (\ref{fig:motivation}, bottom) rather than uniform schemas such as OpenAPI specification. Our results can be summarized as follows: (1) Doc2Agent generated verifiable tools for 59.5\% of real-world APIs and 81.5\% of research APIs, with refinement improving tool pass rates by 47.6\%; (2) compared to direct API calling~\citep{song2025browsingapibasedwebagents}, our tool-based approach achieved a 55.1\% performance boost at only 10\% of the cost per task; (3) we built a glycomaterial research agent using only auto-generated tools, demonstrating Doc2Agent’s potential for scientific applications without domain-specific knowledge.

\section{Doc2Agent Pipeline}
\begin{wrapfigure}{r}{0.5\textwidth}
    \centering
    \includegraphics[width=0.49\textwidth]{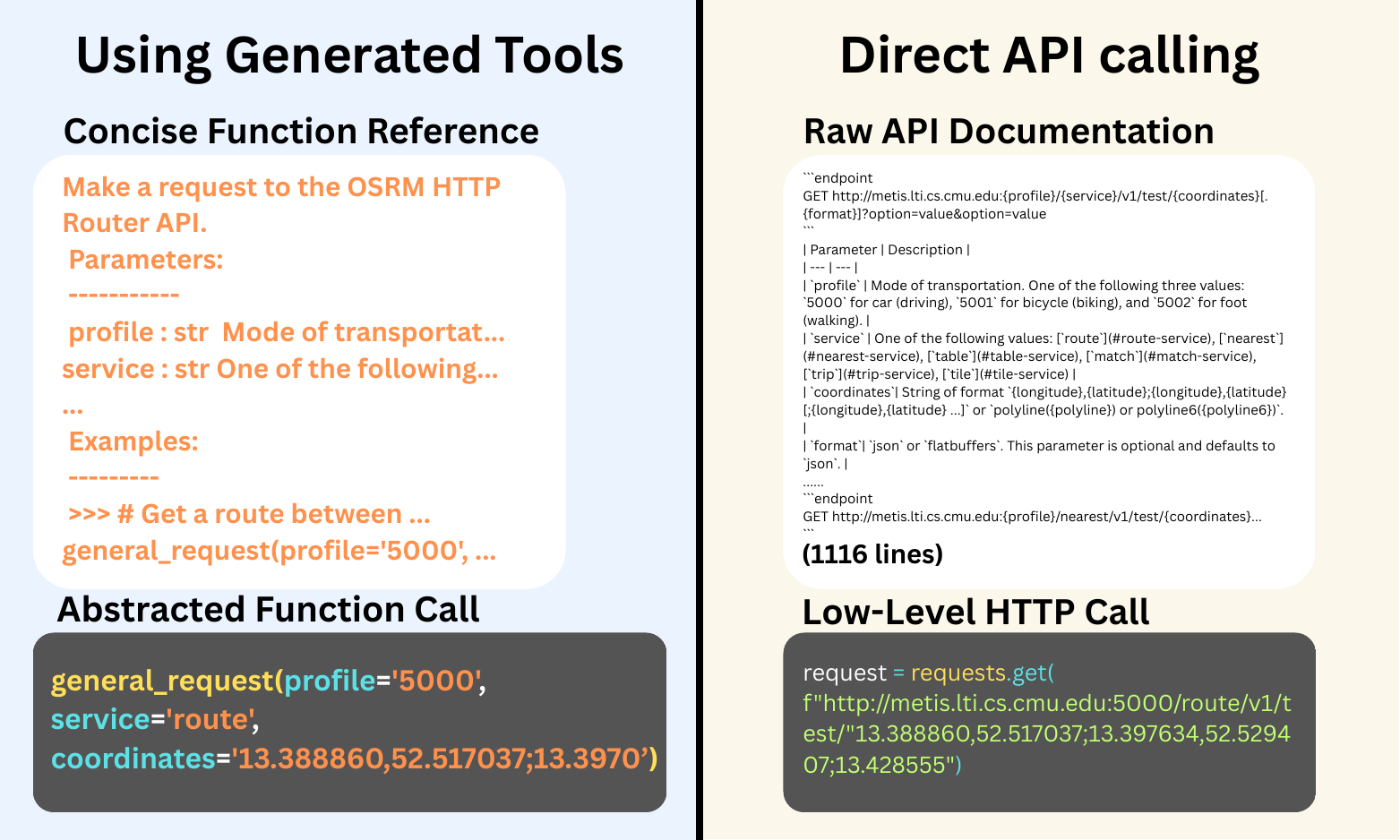}
    \caption{Comparison of function-based API using and direct API calling}
    \label{fig:comparison}
\end{wrapfigure}
\begin{figure*}[tb]
    \centering
    \includegraphics[width=0.9\linewidth]{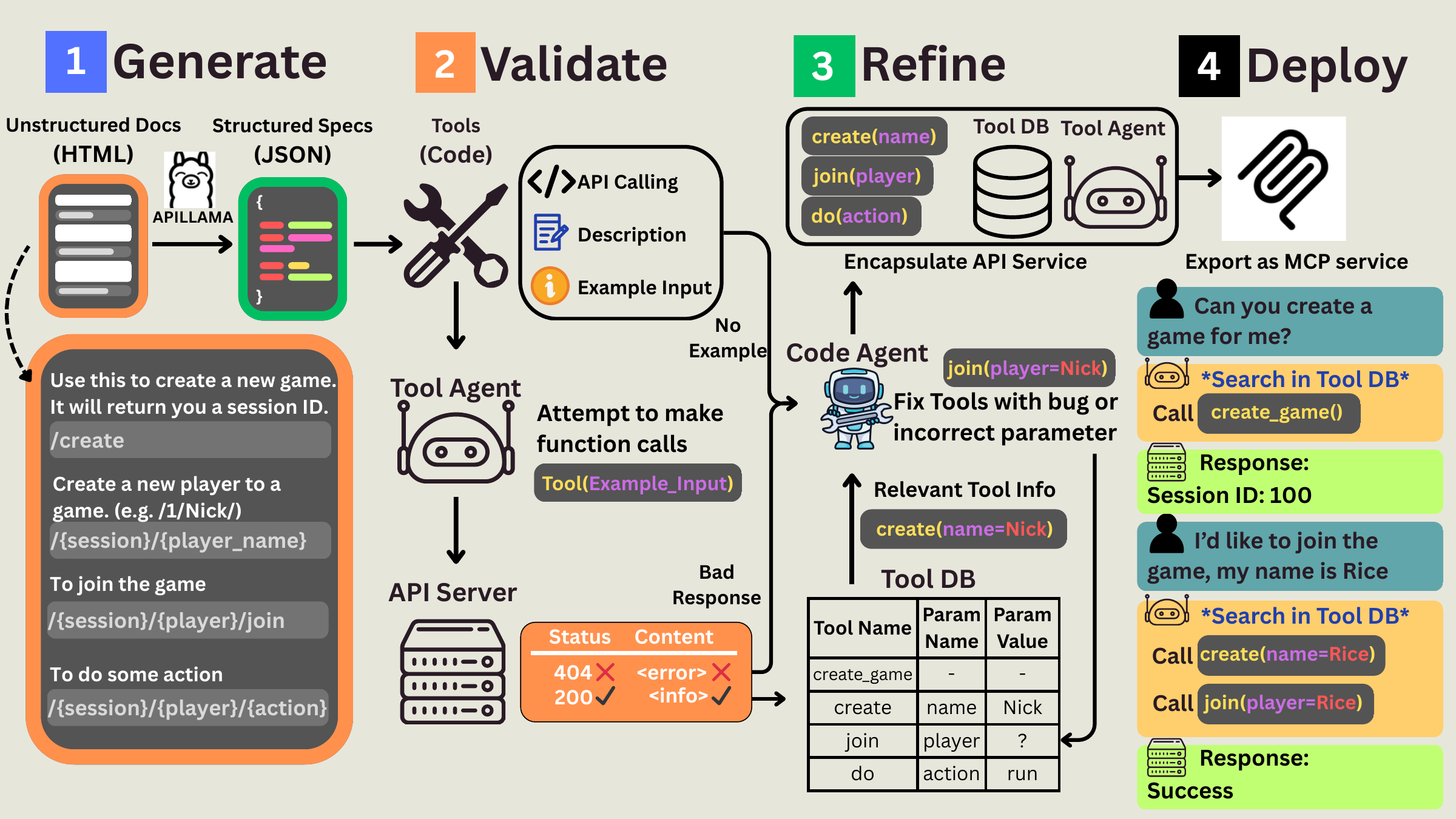}
    \caption{Overview of Doc2Agent: Automated pipeline for generating AI agents from API docs. The API docs in free-text are used to generate tools. The tools are validated and refined. A tool agent equipped with the generated tools is deployed as an MCP service.}
    \label{fig:pipeline}
\end{figure*}

Existing API-calling approaches~\citep{song2025browsingapibasedwebagents} allow LLMs to read API documentation and generate URLs to invoke REST APIs directly. However, this method is cumbersome: lengthy documentation consumes context window space, wastes tokens, and can cause the model to lose focus in multi-turn interactions~\citep{liu2023lostmiddlelanguagemodels, laban2025llmslostmultiturnconversation}.
To address these challenges, we provide agents with generated, Python-based tools in place of raw REST APIs, enabling more efficient and intuitive function calls (Figure~\ref{fig:comparison}). Our workflow can be divided into 4 main steps: tool generation, tool validation, tool improvement, and agent deployment(Figure \ref{fig:pipeline}). We take a set of API documentations in HTML/Markdown format as our input. Each API documentation has multiple endpoints. The output of our pipeline is an agent that can use all API documentations as its tools automatically.
\subsection{Tool Generation}
In tool generation, we produce Python functions with customizable parameters designed for seamless agent use. We adopt two strategies for generating agent-friendly tools:

\textbf{Direct Tool Generation} When API documentation is simple and well structured, we leverage LLMs' structured information extraction capabilities~\citep{dagdelen2024structured, gpt4ostructured} to convert essential API details into standardized JSON format (Appendix~\ref{app:API-extraction-example}). These JSON files are then transformed into Python functions that wrap the API calls, incorporating basic input validation and default value autofill. This encapsulation simplifies API interaction, reduces redundancy, and streamlines agent development.

\textbf{Target-Oriented Tool Generation} For complex or highly flexible APIs, such as general-purpose search endpoints with multiple filters, direct wrapping may be ineffective, as agents struggle to interpret input requirements from the context. In these cases, we guide the generator to first produce simplified function “fingerprints,” which define a specific use case along with expected inputs and outputs. The full function is then generated based on the fingerprint. This strategy produces tools that are better aligned with downstream tasks and more accessible to agents than raw API interfaces.

The test of generated APIs, especially in a real-life environment, may cause unexpected circumstanses, like changing password and deleting items~(Section \ref{sec:limitation}). We allow users to specify REST API methods allowed for tools to minimize the risk of automatic testing.

\subsection{Tool Validation} \label{sec:validation}
The quality of API documentation significantly impacts the performance of API-based agents. However, real-world API documentation is often incomplete or unreliable, making automatic evaluation of generated tools essential.

REST APIs require valid input parameters to function, tools cannot be evaluated without example values. Malformed inputs will fail even if the API itself is functional. When documentation provides example parameters, we assume they produce valid responses and use these tools for evaluation. For tools lacking examples, we infer parameter values in a later stage. 
To validate a tool, we call it using the provided example inputs and compare the actual API response against an expected output generated by a language model, conditioned on the tool’s description. A tool is considered verified if the response aligns with the model-predicted expectation. For tools derived from real-world APIs, our validation results show strong agreement with human judgment. We further analyze failed tools using status codes, response content, and runtime exceptions. Most failures stem from incorrect or incomplete parameter values, motivating our method for automatic generation of high-quality parameter inputs(Appendix \ref{app:error-validation}). 

\subsection{Tool Refinement}
\textbf{Parameter Value Inference} \label{sec:API-KB}
\label{sec:example-parameter-generation}
For APIs lacking proper parameter values, large language models often struggle to generate valid inputs. Our experiments (Section \ref{glycogpt}) confirm that LLMs are unreliable at guessing parameter values. This aligns with recent findings~\citep{song2025callnavichallengeempiricalstudy}, which highlight the difficulty LLMs face in producing complete and accurate API inputs. We construct an automatically generated parameter database to support parameter value inference. We leverage two primary sources of information: (1) parameter examples from other API documentations, particularly from the same domain, and (2) JSON responses from previously validated API tools, which contain rich domain-specific key-value pairs. The latter captures implicit inter-API dependencies, where outputs of one service often correspond to inputs of another—conceptually forming a service dependency graph~(\citet{bushong2021microservice, Lercher_2024}). 
These dependencies allow us to repurpose response data to infer plausible parameter values.
The database is constructed without requiring external domain knowledge. All discovered parameter values are stored in a vector database to enable efficient semantic similarity search~\citep{han2023comprehensivesurveyvectordatabase}. 
\begin{wrapfigure}{r}{0.6\textwidth}
\centering
\includegraphics[width=0.57\textwidth]{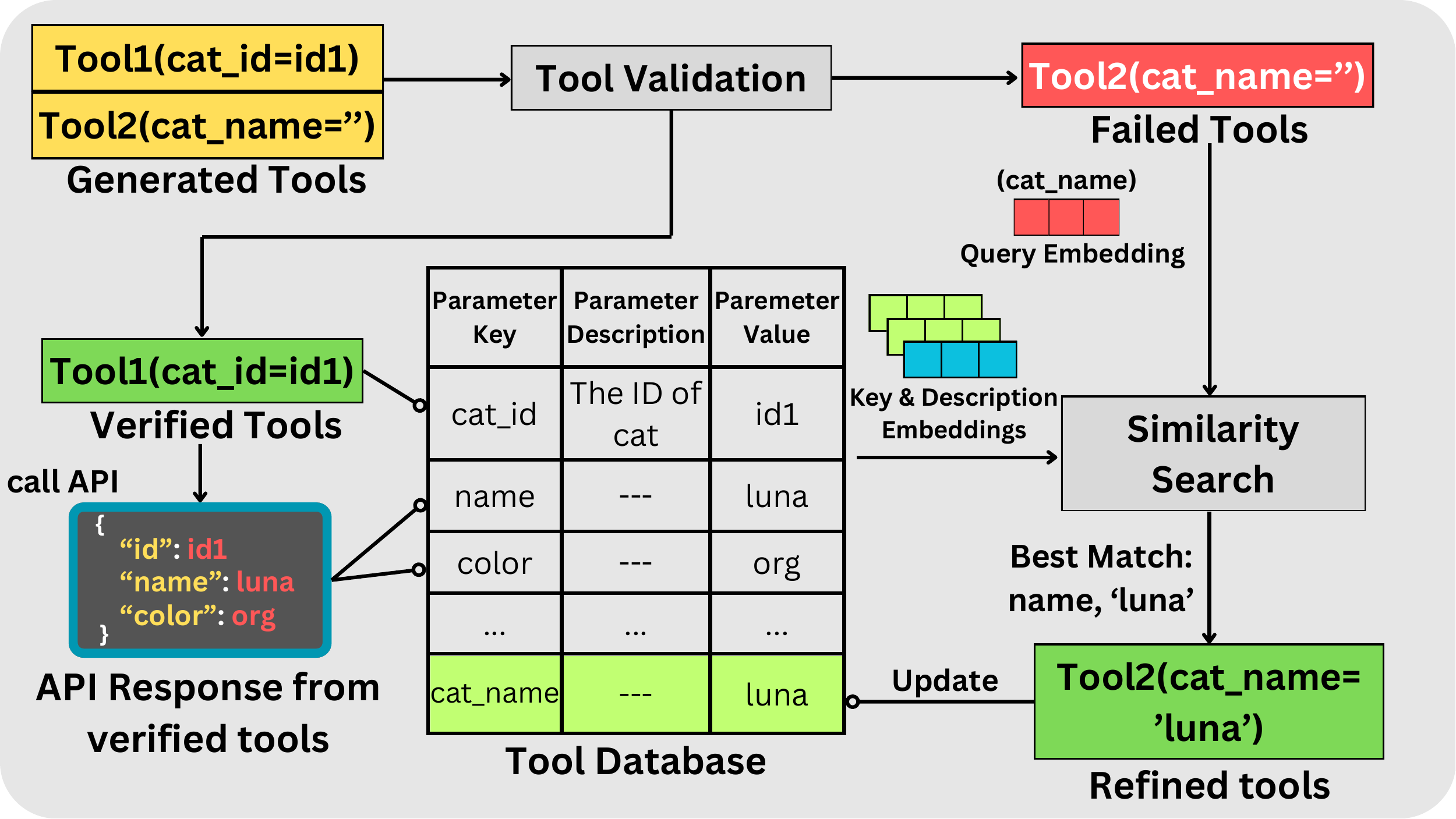}
\caption{A parameter database is automatically constructed using validated tools, enabling parameter value inference based on the semantic similarity of parameter/response keys.}
\label{fig:parameter-value-inference}
\end{wrapfigure}
Parameter examples are indexed by both name and description. When encountering an unknown parameter $\gamma$, we embed its name and description, retrieve semantically similar entries from the database, and rank them based on tool and parameter similarity. We then sample up to 10 candidate values for downstream usage. 

\textbf{Tool Fixing} We fix the tools that either have no parameter values or produced incorrect responses. For each failed tool, we provide a code agent with the corresponding API documentation, error information, and the original Python code. If the failure stems from missing parameters, we additionally supply the agent with candidate parameter values.

The agent generates a revised version of the tool, which is then re-evaluated through the validation process. If the updated tool passes validation, we overwrite its code and documentation. When a new parameter value is used, it is recorded as an example and added to the parameter database. This \textbf{refinement-validation loop} is repeated for multiple rounds to maximize tool recovery and quality.

\subsection{Deployment}
Once high-quality tools are obtained, they can be readily integrated into tool-using agents such as CodeAct~\citep{wang2024executablecodeactionselicit}. Since the tools are implemented in Python, they remain compatible with a wide range of agent architectures, including those that support advanced capabilities like planning~\citep{stein2025automatinggenerationpromptsllmbased}, reasoning(\citet{wei2023chainofthoughtpromptingelicitsreasoning}, \citet{kojima2023largelanguagemodelszeroshot}), long-term memory~\citep{du2025rethinkingmemoryaitaxonomy}, and tool retrieval~\citep{yehudai2025surveyevaluationllmbasedagents}. This synergy between agents and tools significantly enhances task-solving performance.

To support scalable and standardized integration, tools can be deployed via an MCP (Model-Context-Protocol) server~\citep{hou2025modelcontextprotocolmcp}, which provides a unified protocol for accessing tools, managing data, and orchestrating workflows. Tools can operate independently through a host-side agent or be dynamically invoked by clients, enabling flexible and modular agent-tool interactions. We deploy the API services using \textit{FastAPI MCP}~\citep{fastmcp}. As an alternative deployment path, tools can also be exported as OpenAPI specifications~\citep{openapi}, enabling seamless integration with both enterprise systems (e.g., \href{https://openai.com/index/introducing-gpts/}{GPTs}) and open-source agent frameworks such as \href{https://www.crewai.com/}{CrewAI}, \href{https://www.langchain.com/langgraph}{LangGraph}, and \href{https://microsoft.github.io/autogen/dev//index.html}{AutoGen}.

\section{Experiment}
\subsection{Data Collection}
Previous REST API-based benchmarks (\citet{zhou2024webarenarealisticwebenvironment}, \citet{trivedi2024appworldcontrollableworldapps}, \citet{zhong2025complexfuncbenchexploringmultistepconstrained}) typically assume ideal conditions by either providing agent-ready, Python-based tools or clean, standardized API specifications. In contrast, our work emphasizes \textbf{robustness and scalability} in the face of real-world API documentation, which is often unstructured, inconsistent, and heterogeneous in format and quality. To evaluate the effectiveness of our pipeline under these realistic conditions, we curated data from three diverse sources.

\noindent \textbf{Real-world API documentations} We collect 167 API documentation pages from \href{https://apilist.com}{APIList.com}, comprising a total of 744 endpoints. Upon analysis, we find that only 24 of these documentations were of high quality. The majority were semi-structured and difficult to interpret at first glance, often requiring careful manual inspection to infer correct parameter values. Examples and selection criteria are provided in Appendix \ref{app:real-life-api}.

\noindent \textbf{WebArena}~\citep{zhou2024webarenarealisticwebenvironment} WebArena is a reproducible web environment designed for developing autonomous agents to perform complex web-based tasks. While originally intended for web browsing agents, its environment also supports API interactions, with API documentation quality varying across domains~\citep{song2025browsingapibasedwebagents}, which provides a great environment for testing robust tool generation approach. We utilize the available API documentation from five WebArena environments: GitLab, Open Street Map(Map), One Stop Shop(Shopping-Customer), E-Commerce Content Management(Shopping-Admin, CMS) and Cross site tasks(Multi). We exclude the Reddit task, because WebArena Reddit sandbox doesn't provide API documentations.

\noindent \textbf{Glycoscience APIs} To evaluate our pipeline’s ability to generate domain-specific research agents, we target glycoscience~\citep{f91d0fc0f6f944c0a108d450c6aa7490}, with complex APIs and inconsistent data standards. We collect REST API documentation from major databases, including \textbf{GlycoData}~\citep{glylcodata}, \textbf{GlyGen}~\citep{york2020glygen}, \textbf{GlyTouCan}~\citep{tiemeyer2017glytoucan}, \textbf{KEGG GLYCAN}~\citep{hashimoto2006kegg}, \textbf{Glycosmos}~\citep{yamada2020glycosmos}, \textbf{Glyconnect}~\citep{alocci2018glyconnect}, \textbf{The O-GlcNAc Database}~\citep{wulff2021human}, \textbf{GLYCAM}~\citep{Glycam}, \textbf{Protein API}~\citep{10.1093/nar/gkx237}, \textbf{PubChem}~\citep{kim2016pubchem} and \textbf{UniLectin}~\citep{imberty2021unilectin}.. These APIs vary widely in quality and structure, and present additional challenges such as inconsistent identifiers, non-standard representations, and cross-database linking (Appendix \ref{app:glyco-api-complexity}).

\subsection{Implementation}
\textbf{Generator} We use \verb|GPT-4o| in structured mode for direct tool generation and \verb|Claude 3.7 Sonnet| for target-oriented generation. In the direct tool generation, one tool is generated per extracted API. In the target-oriented generation, up to 10 task-specific tools are generated per API documentation. For WebArena, we use direct tool generation for Gitlab and Map docs; and task-oriented generation for Shopping and Admin docs.

\noindent \textbf{Code Agent for Validation and Refinement} We employ \verb|Claude 3.7 Sonnet| as the code agent for tool testing and refinement. It is capable of generating well-structured Python code along with clear documentation, enabling seamless integration with downstream agents.

\noindent \textbf{Tool Agent} To evaluate the impact of high-quality, agent-usable tools, we conduct a comparative study based on the design of API agents from \citet{song2025browsingapibasedwebagents}. Our implementation builds upon CodeAct \cite{wang2024executablecodeactionselicit}, powered by GPT-4o, with basic code execution capabilities and minimal tool retrieval support. The key difference lies in tool usage: our API agent utilizes refined, Python-based API tools, whereas prior approaches rely solely on raw API documentation as input prompts. For more advanced applications, stronger tool agents such as ReTool \cite{feng2025retoolreinforcementlearningstrategic} can be used to further enhance tool utilization.

\section{Result}
\subsection{Generation of Python-based API Tools from Documentation} \label{sec:experiment-2}
Our validator enables self-evaluation of the generated tools. An example of the generated tool is provided in Appendix~\ref{app:tool-example}. We allow up to three refinement rounds per tool. Results are summarized in Table~\ref{tab:experiment-validate-tool}. Despite the complexity of real-world API documentation, we successfully produce 443 validated tools. A frequent cause of failure is the REST API service no longer operational. In the WebArena benchmark, Wiki and Map environments have relatively simple API designs, and we apply direct tool generation. Some Map APIs fail due to authentication errors, which is not required in WebArena task. In contrast, Shopping-Admin, Shopping-Customer and GitLab involve complex API usage with numerous endpoints. For these, we use both direct tool generation and target-oriented tool generation to produce more task-aligned, agent-friendly tools. For Glycoscience APIs, most tools pass validation without refinement, as they primarily consist of information retrieval endpoints.

\begin{table}[tb]
\resizebox{\linewidth}{!}{
\begin{tabular}{lccccccc}
\toprule
\multirow{2}{*}{} & \multirow{2}{*}{\textbf{Real-life API}} & \multicolumn{5}{c}{\textbf{WebArena}}                             & \multirow{2}{*}{\textbf{Glycoscience API}} \\ \cmidrule(rl){3-7}
                     &                                & Wiki & Map & Shopping-Admin(CMS) & Shopping-Customer & GitLab &                                    \\ \midrule
Endpoints        & 744                            & 26   & 53  & 555            & 108               & 988    & 131                                \\
Validated Tools      & 443                            & 21   & 28  & 159            & 35                & 213    & 70  \\ \bottomrule                             
\end{tabular}
}

\caption{Generated Agent-Ready Python-based Tools with Doc2Agent from Raw API Docs}
\label{tab:experiment-validate-tool}
\end{table}
 
\subsection{Tool agent performance on WebArena} \label{sec:webarena-experiment}
To compare our tool-using agent with direct API-based agent~\citep{song2025browsingapibasedwebagents}, we keep most components of the CodeAct agent unchanged, including tool retrieval, planning, and memory. As shown in Table~\ref{tab:experiment-api-compare}, our method outperforms direct API calling with a 55\% relative increase in average success rate. Notably, Doc2Agent achieves a 57.8\% improvement on the Shopping task and a 95.6\% improvement on the CMS task, both of which provide low-quality API documentation. These results highlight the effectiveness of a tool-based approach over direct API calls, especially when documentation is incomplete or poorly structured. Even for well-documented APIs like Map, performance improves by 29.3\% due to simplified parameter usage. The substantial overall performance gain enables our pure tool-based agent to outperform the hybrid approach that combines direct API calls with browser interactions. Meanwhile, repeated use of tools significantly reduces token consumption. On average, our approach costs \$0.12 per task, compared to \$1.20 for direct API calling and \$1.50 for the hybrid approach.

\begin{table*}[tb]
\resizebox{\linewidth}{!}{
\begin{tabular}{lcccccc}
\toprule
\textbf{Method}        & \textbf{Gitlab} &\textbf{Shopping}  & \textbf{CMS}  & \textbf{Map} & \textbf{Multi} &\textbf{Avg.}\\ \midrule
\rowcolor{gray!20}  \multicolumn{7}{c}{Vanilla WebArena Evaluation} \\ 
WebArena Baseline$^{\color{blue} \textbf{b}}$\citep{zhou2024webarenarealisticwebenvironment} &15.0 &13.9 &10.4 &15.6 &(8.3) &(12.3) \\
SteP$^{\color{blue} \textbf{b}}$\citep{sodhi2024stepstackedllmpolicies} &32.2 &50.8 &23.6 &31.2 &(10.4) &(36.5) \\ 
SkillWeaver$^{{\color{blue} \textbf{b}}}$\citep{zheng2025skillweaverwebagentsselfimprove} &22.2 &27.2 &25.8 &33.9 &- &(29.8)\\\midrule
API-based agent$^{\color{red} \textbf{t}}$\citep{song2025browsingapibasedwebagents}       &43.9 &25.1   & 20.3           & 45.4        & (8.3)  &(29.2)\\ 
Hybrid agent$^{{\color{blue} \textbf{b}} {\color{red} \textbf{t}}}$\citep{song2025browsingapibasedwebagents} &44.4 &25.7 &41.2 & 45.9 & (16.7) &(38.9)\\ \midrule
\rowcolor{gray!20}  \multicolumn{7}{c}{API-Specified Evaluation} \\
Hybrid agent$^{{\color{blue} \textbf{b}} {\color{red} \textbf{t}}}$\citep{song2025browsingapibasedwebagents} &47.2 &29.4 &\textbf{45.5} &50.5 &\textbf{44.0} &42.2\\ \midrule
\makecell[l]{\textbf{Doc2Agent}$^{{\color{red} \textbf{t}}}$\textbf{(Ours)}}         &\makecell{\textbf{48.9}}   &\makecell{\textbf{39.6}}  & \makecell{39.7}           & \makecell{\textbf{58.7}}    &\makecell{$\textbf{44.4}$}   &
\makecell{$\textbf{45.3}$}\\ 
$\Delta$ vs Direct API Calling & \colorcell{11.4} & \colorcell{57.8} & \colorcell{95.6} & \colorcell{29.3} &- &\colorcell{55.1}\\
\bottomrule
\end{tabular}
}
\caption{Comparison result of different Methods on WebArena. We exclude Reddit task due to no API documentation. Numbers in parentheses include Reddit tasks. $\Delta$ represents the estimated relative improvement percentage. $^{\color{blue} \textbf{b}}$ Browser-based Agent $^{\color{red} \textbf{t}}$ API-based Agent}
\label{tab:experiment-api-compare}
\end{table*}

In comparison, SteP~\citep{sodhi2024stepstackedllmpolicies} relies on manually defined policies and task-specific prompts to guide agent actions. Our Doc2Agent outperforms SteP on all tasks except Shopping, demonstrating the superior effectiveness of automatically generated tools. SkillWeaver~\citep{zheng2025skillweaverwebagentsselfimprove} synthesizes reusable skills as Python-based browser tools, which are relatively inefficient compared to our generated API-based tools. Doc2Agent consistently achieves higher performance across all tasks compared to SkillWeaver.

\textbf{API-Specified Evaluation} The original WebArena evaluator is tailored for browsing-based agents, relying on string matching, URL matching, and site navigation. We observed several false positives, such as cases where the agent failed to complete the task but included partial or coincidental keywords. Additionally, evaluation criteria involving browser interactions (e.g., editing web elements) are not applicable to API-only agents. To address these limitations, we introduce two adjustments: (1) For tasks marked as successful by the WebArena evaluator, we use an LLM to verify whether the task was genuinely completed or merely matched keywords by chance. (2) We restructure site navigation and content-checking evaluations into \verb|exact_match| or \verb|must_include| criteria, based on the agent’s final action log. This enables direct assessment of agent outputs, retrieved content, or API JSON responses, ensuring fair evaluation for API-based agents. We apply this API-specific evaluation to the hybrid agent as well, observing consistent overall performance with minor improvements. We were unable to re-evaluate the API-based agent from prior work due to the lack of publicly available logs. The evaluation details are provided in Appendix~\ref{app:evluation}.

\subsubsection{Discussion: What factors make good tools for agents?} 
\textbf{+ Accurate Parameter Values} Our pipeline is particularly effective in addressing incomplete or vague API documentation. The code agent infers valid input parameter values by interacting with the environment—an approach especially beneficial for APIs like Shopping-Customer and GitLab, which return user-specific data not described in the documentation. By first querying supportive endpoints (e.g., \verb|list_project|), the agent gathers contextual information to validate additional tools. We observed a strong negative correlation between parameter complexity and success rate for HybridAgent on Map tasks, highlighting the challenge of manually specifying complex inputs. In contrast, tools generated by Doc2Agent tend to use parameters with higher semantic alignment and lower complexity. However, validating parameters for stateful REST methods remains difficult (see Section~\ref{sec:limitation}) and introduces safety concerns for automatic testing on public servers.

\textbf{+ Function-usage Success Rate} Function usage success rate has the highest correlation with the task-level success rate. Doc2Agent calls functions consistently well(85\%-97\%) across all sites, while HybridAgent excels in shopping and CMS(98\%-99\%) tasks. 

\textbf{- Complex Response JSON} Our approach shows lower performance than the Hybrid agent on CMS tasks, primarily due to the presence of long and unfiltered API responses. To manage context limitations, we truncated the response content, which may cause information loss. This highlights the need for more sophisticated response processing. Integrating advanced JSON navigation or summarization techniques could significantly improve the effectiveness of API-based agents in such settings. Moreover, developing  tools for response-side information filtering presents a promising direction for enhancing tool efficiency.


\subsection{Generation and Deployment of Research Agent} \label{glycogpt}


Unlike many real-world APIs, research-domain APIs often involve complex database identifiers and representations (see Appendix~\ref{app:tab:glycodataset}), along with relatively limited domain knowledge in base model. In this experiment, we explore the applicability of Doc2Agent in a challenging scientific setting, where we apply Doc2Agent to glycoscience APIs and automatically constructed a tool-using research agent specialized for the glycan domain.

\textbf{Tool-generation} Using Doc2Agent, we generate 70 refined tools and automatically constructed a database of parameter name–value pairs through parameter value inference. We compare two methods for parameter value acquisition: an API response–based inference approach and GPT-generated parameter candidates. Our results show that the API-based inference approach doubles the tool pass rate (see Appendix~\ref{app:parameter-value-infernence-comparison}).

\textbf{Settings} To enable seamless access to the research agent, we host the tools on an MCP server using FastMCP~\citep{fastmcp}. We evaluate three settings: (1) As used in Section~\ref{sec:webarena-experiment}, we apply CodeAct which loads tools directly into functions (2) An entirely open-source setup using \verb|Qwen-Agent|~\citep{qwen-agent} as the agentic framework, which supports basic MCP tool orchestration, with \verb|Qwen3-32B|~\citep{yang2025qwen3technicalreport} as the base model.
(3) A deployment using the enterprise-level application \href{[https://claude.ai/download}{Claude Desktop}, which offers advanced MCP Connector integration with \href{[https://www.anthropic.com/news/claude-4}{Claude Sonnet 4} as the base model. 

\textbf{Task Generation} We prompted GPT-4o to generate 50 research-oriented tasks based on the tool descriptions, including 30 single-tool tasks and 20 multi-tool tasks. Due to the lack of a formal evaluation framework, we used an LLM-as-a-judge approach (see template in Appendix~\ref{app:llm-as-a-judge-prompt}) to estimate task success. Since not all generated tasks are guaranteed to be solvable, we report success rates over the Filtered set using the union of all tasks successfully completed by at least one agent to provide a fair basis for relative performance comparison.

\textbf{Results} Table~\ref{tab:glyco-agent} presents our results. Among the tested agents, Claude achieved the highest average success rate of 36\%, successfully solving 58.1\% of the do-able tasks. Claude excelled in multi-tool tasks, leveraging its advanced MCP orchestration technique. CodeAct's success rate is comparable to its performance on WebArena tasks, suggesting that our function generation method is not constrained by domain-specific characteristics and has the potential to scale effectively to other research domains. Notably, Qwen3-32B, despite being a smaller base model, performed on par with CodeAct, underscoring the potential of open-source research-tool agents. 

\begin{table}[htbp]
\resizebox{\linewidth}{!}{
\begin{tabular}{lccccccc}
\toprule 
 \multirow{2}{*}{\textbf{Setting}}         & \multirow{2}{*}{\textbf{Total}} & \multirow{2}{*}{\textbf{Filtered}}         & \multicolumn{2}{c}{\textbf{Tool Use}} & \multicolumn{3}{c}{\textbf{Task Type}}              \\ \cmidrule(lr){4-5} \cmidrule(lr){6-8}
          &  &  & Single          & Multi         & Analysis & Data Retrieval & Transformation \\ \midrule
\textbf{Claude Desktop(Claude-Sonnet-4)}    & \textbf{36.0}  & \textbf{58.1}                      & 36.7            & \textbf{35.0}          & \textbf{45.5}     & \textbf{30.4}           & 20.0           \\
\textbf{CodeAct(GPT-4o)}   & 32.0  & 51.6                      & \textbf{46.7}            & 10.0          & 40.9     & \textbf{30.4}           & 0.0            \\
\textbf{QwenAgent(Qwen3-32B)} & 30.0  & 48.3                      & 33.3            & 25.0          & 31.8     & 26.1           & \textbf{40.0}          \\ \bottomrule
\end{tabular}
}
\label{tab:glyco-agent}
\caption{Success rate over different agent frameworks. \textbf{Filtered} refers to evaluation on the combined set of tasks successfully completed by at least one framework.}
\end{table}



\section{Related Works}
\textbf{Tool(API) Agent} LLM-based tool agents are able to reason through user queries, select and apply appropriate actions, and return the results of the chosen action. For example, \citet{bran2023chemcrowaugmentinglargelanguagemodels} developed ChemCrow by integrating GPT-4 and 18 tools designed by experts. \citet{qin2023toolllmfacilitatinglargelanguage} collected 16k public REST APIs from the RapidAPI platform \cite{rapidapi}, and trained a tool retriever that can choose the most appropriate API in response to a user query. \citet{wang2025toolgenunifiedtoolretrieval} represented tools as a unique token that are integrated into LLM generation. \citet{zhang2025apiagentsvsgui} compared API agents and GUI agents in solving web tasks. \citet{song2025browsingapibasedwebagents} built a hybrid agent that can browse and call APIs to perform online tasks. Model Context Protocol\cite{hou2025modelcontextprotocolmcp} allows fast deployment of argentic services. Benchmarks\cite{yehudai2025surveyevaluationllmbasedagents} are developed to evaluate the effectiveness of agents, while most of current benchmarks assume access to well-prepared toolsets(\citet{trivedi2024appworldcontrollableworldapps}, \citet{qin2023toolllmfacilitatinglargelanguage}, \citet{song2023restgptconnectinglargelanguage}). Particularly, we use WebArena\cite{zhou2024webarenarealisticwebenvironment} to demonstrate the improvement by encapsulated API-based tools.

\textbf{Agent Generation and Improvement} The creation for AI agents requires prompt design and tool design, which will be challenging to be automated. \citet{chen2024autoagentsframeworkautomaticagent} proposed a framework that generates specialized agents to form an AI team tailored to specific tasks. \citet{shi2025toollearningwildempowering} proposed AutoTools, which explores function generation through docs, but relies on standardized API specifications from RapidAPI. \citet{zheng2025skillweaverwebagentsselfimprove} introduced Skillweaver, a framework that enables web agents to self-improve by practicing reusable skills into APIs. However, their test revealed synthesized skills are worse than human APIs. Another direction(\citet{gutiérrez2025ragmemorynonparametriccontinual}, \citet{du2025rethinkingmemoryaitaxonomy}) focus on memory updating that allows agents to improve through past experience.

\section{Conclusion}
In this work, we present \textbf{Doc2Agent}, a scalable pipeline for generating AI agents equipped with validated, Python-based tools from natural language REST API documentation. Doc2Agent not only converts unstructured documentation into executable tools but also detects and refines inaccuracies through automated validation. We applied our pipeline to three sources: real-world APIs, WebArena environments, and glycomaterial research APIs. We successfully generated and validated a large set of agent-ready tools. In WebArena, agents using our tools achieved a 55\% relative performance improvement while reducing per-task cost to just 10\% of the original. We further demonstrated Doc2Agent’s domain adaptability by building a research agent for glycomaterial science, leveraging our parameter inference method to resolve diverse data representations without external expert input. Doc2Agent enables efficient and robust agent creation across domains, bridging the gap between unstructured API documentation and practical tool-based agent deployment.

\section{Limitation} \label{sec:limitation}
\textbf{API Documentation Acquisition} Our method relies on available API documentation (e.g., web pages, OAS files) to generate Python-based tools, and thus cannot be applied to APIs lacking documentation. Currently, some human effort is still required to collect the initial API documentation. A promising future direction is to integrate our approach with web browsing capabilities, enabling agents to automatically discover and scrape relevant API documentation based on a given task—thereby expanding the agent’s toolset autonomously.

\textbf{Validation for Stateful APIs} Our current validation approach relies on analyzing individual API responses, which is effective for stateless APIs but insufficient for stateful ones. Validating stateful APIs often requires more complex workflows involving multiple, coordinated API calls and additional endpoints to query server-side status or track changes over time.

\textbf{API dependency} We infer values for unknown parameters using example inputs and JSON responses from validated API calls. This approach leverages implicit dependencies between APIs and performs well when such relationships exist. However, its effectiveness diminishes for unrelated APIs, where parameter values cannot be reliably inferred from prior tool outputs.

\textbf{Evaluation} Evaluating our pipeline poses unique challenges. Existing function-calling benchmarks (e.g., ToolBench~\citep{qin2023toolllmfacilitatinglargelanguage}, AppWorld~\citep{trivedi2024appworldcontrollableworldapps}, ComplexFuncBench~\citep{zhong2025complexfuncbenchexploringmultistepconstrained}) focus primarily on tool usage, assuming well-prepared APIs and thus bypassing the need for tool generation. In contrast, browsing-based benchmarks (e.g., WebArena~\citep{zhou2024webarenarealisticwebenvironment}, WebVoyager~\citep{he2024webvoyagerbuildingendtoendweb}, ST-WebAgentBench~\citep{levy2024stwebagentbenchbenchmarkevaluatingsafety}) introduce biases when used to evaluate our method: not all tasks in these environments are solvable via API calls, while others are directly derived from API usage. Consequently, comparisons between API agents and browsing agents on these benchmarks may be skewed due to task distribution. This highlights the need for a benchmark specifically designed to evaluate \textbf{open-domain API usage}, where tool generation, validation, and application can be fairly assessed across diverse and realistic API scenarios.

\textbf{Cheating in Code Agent} We observed that the code agent occasionally attempted to "cheat" during tool refinement. For example, by generating \verb|try-catch| blocks to suppress exceptions and bypass validation errors. To mitigate this issue, we introduced an uneditable testing code section to enforce strict validation. However, this behavior highlights a broader concern: the agent's optimization strategy may not always align with human intent or practical utility.

\section{Ethical Consideration}
Autonomous REST API agents may trigger unintended or harmful actions, especially when interacting with external services\cite{mudryi2025hiddendangersbrowsingai}. The tools generated by our pipeline are not manually reviewed for safety and, if misused, could result in undesirable or potentially malicious behavior. Given the ongoing development of AI safety practices, we recommend restricting tool generation to \verb|GET| methods only, which are generally read-only and pose lower risk. This precaution helps mitigate unintended side effects during agent execution.

\bibliography{colm2025_conference}
\bibliographystyle{colm2025_conference}

\appendix
\renewcommand{\thetable}{\Alph{section}\arabic{table}}
\setcounter{table}{0}
\renewcommand\thefigure{\thesection\arabic{figure}}    
\setcounter{figure}{0}

\section{Real-life API Dataset} \label{app:real-life-api}
\subsection{Constuction}
\textbf{Documentation styles} The API documentation we collected can be categorized into three levels based on the organization and clarity of the API descriptions: (1) \textbf{Fully organized} The documentation follows a well-defined template, providing all necessary information to call the API in a structured and comprehensive way. Use cases are clearly explained, often with example code. API documentation on platforms like RapidAPI Hub and Postman API typically fall into this category. (2) \textbf{Semi-organized} This type of documentation includes basic descriptions but lacks clarity for each endpoint. Some essential information may not be labeled with specific keywords and is instead embedded within general text. Additional effort is often required to identify key details. (3) \textbf{Unorganized} These documents are minimal, often missing example code or detailed descriptions. They require some level of inference and reasoning to understand the API's usage, with clues only available through endpoint names. We show that our benchmark consists of mostly semi-structure documentations, and only a few documentations are fully organized(Appendix \ref{app:api-doc-classification}).

\textbf{Selection criteria} We filtered for API documents that do not require API keys, allowing for easier and more convenient API access without needing authentication, which often involves submitting forms or linking payment methods. Although automating API key sign-up is feasible using web agents \cite{autosignup}, we opted for APIs without authentication requirements to streamline the process. In total, 347 unique API documents were selected and downloaded in HTML format. Since some links pointed to index pages or API information that was dynamically loaded via JavaScript, we employed a large language model to identify pages containing static HTML code with API endpoints. This approach ensured that we captured only the documentation with accessible and actionable API details.

Due to variations in the quality and completeness of API documentation, we extracted only the essential information needed for tool generation. Specifically, for each API documentation, we captured the \verb|base URL| and a list of endpoints. For each endpoint, we extracted the \verb|endpoint path|, \verb|required parameters|, \verb|optional parameters|, and a brief \verb|description|. In cases where the base URL was not specified, human annotation was necessary. The schema used for this extraction is provided in the Appendix \ref{app:json-schema}.

To implement this extraction, we defined the schema using a Pydantic model and employed GPT-4o in structured mode to parse the HTML documents and extract the desired information. After filtering out pages that lack API information (primarily product index pages), we obtained 167 API documentation with 744 endpoints and extracted their structured information in JSON format. An example of input API documentation and output JSON structure is in Appendix \ref{app:API-extraction-example}.
\subsection{Documentation Quality Classification} \label{app:api-doc-classification}
\begin{figure}[htbp]
\centering
\subfigure[]{
\includegraphics[width=0.5\linewidth]{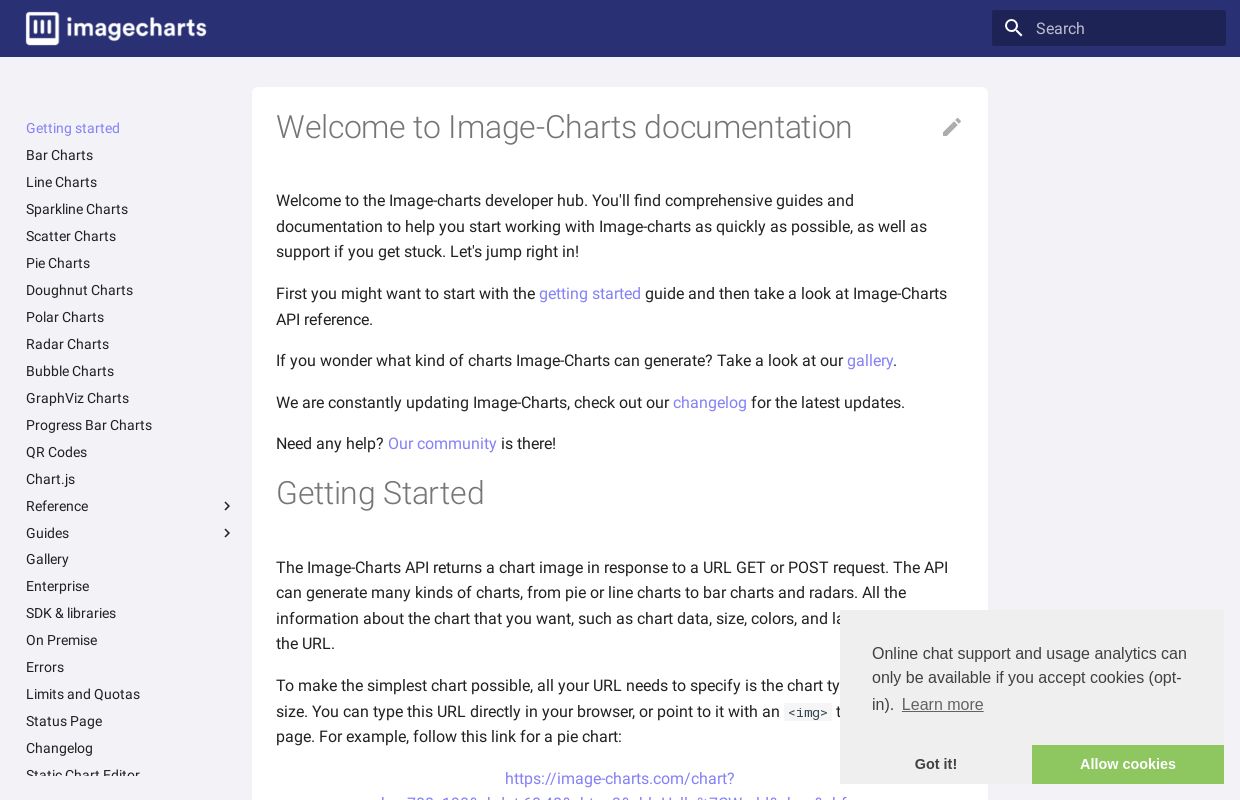}
}
\subfigure[]{
\includegraphics[width=0.5\linewidth]{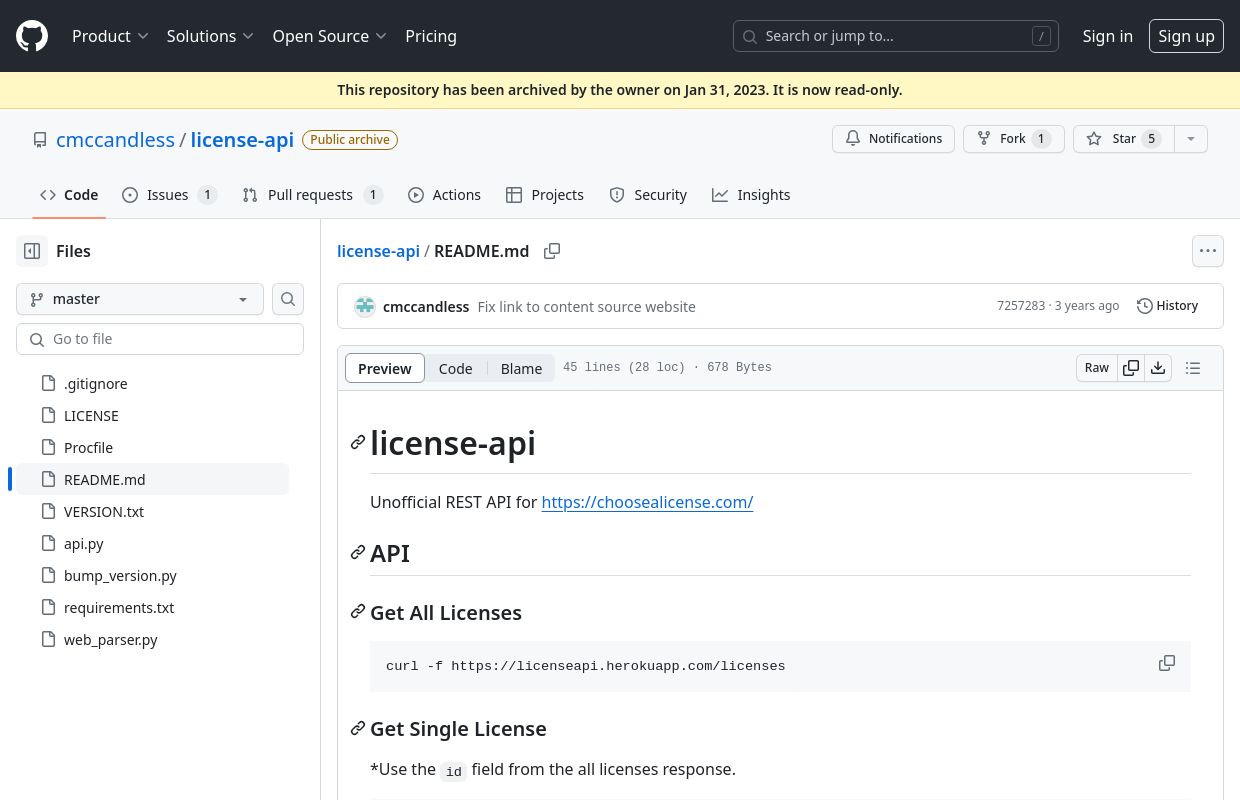}
}
\subfigure[]{
\includegraphics[width=0.5\linewidth]{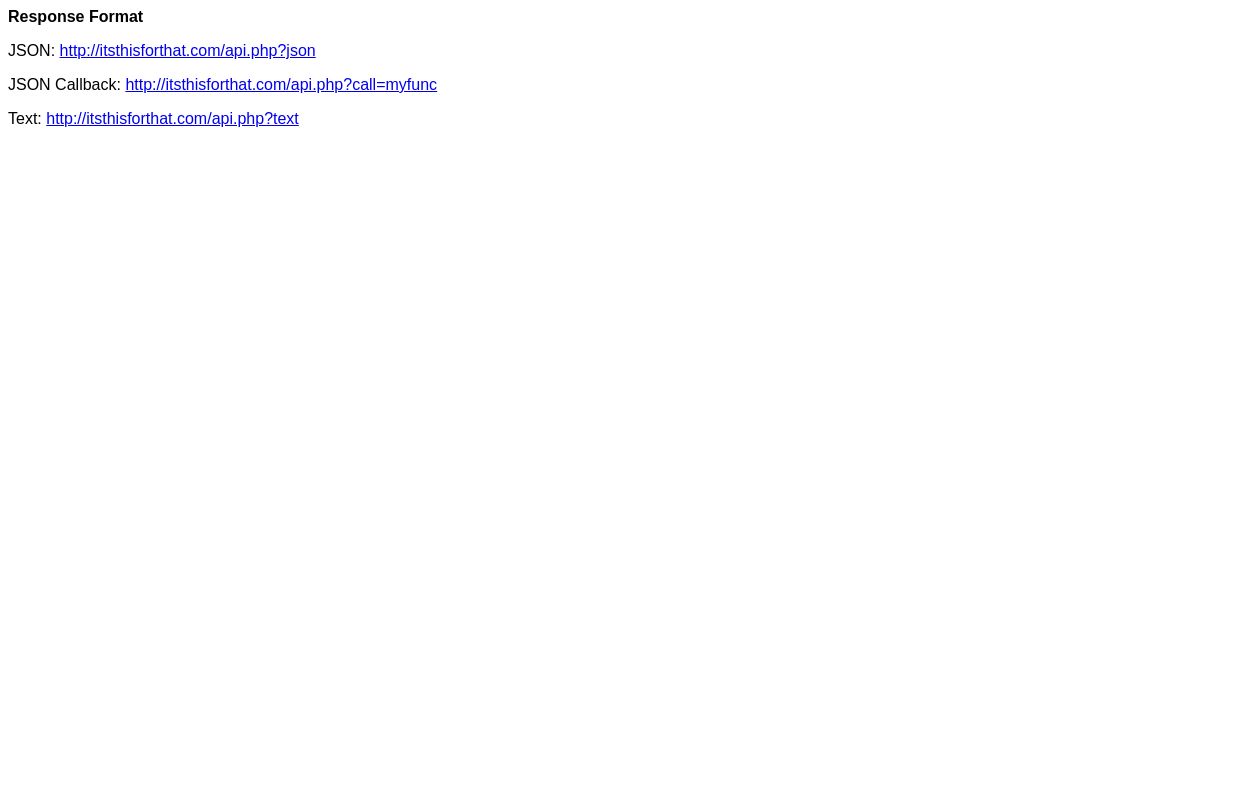}
}
\caption{Example of API documentation of each category. (a) is an example of organized API documentation \href{https://documentation.image-charts.com/?utm_source=apislist.com}{\textbf{ link}}, (b) is the example of semi-organized documentation \href{https://github.com/cmccandless/license-api/blob/master/README.md?utm_source=apislist.com}{\textbf{ link}}, and (c) is the unorganized API documentation \href{https://itsthisforthat.com/api.php?utm_source=apislist.com}{\textbf{ link}}}
\label{app:fig:api-webpage-example}
\end{figure}

We classify API documentations into 3 categories: \textbf{Fully organized}, \textbf{Semi-organized} and \textbf{Unorganized}, based on their clarity and completeness of information. Using GPT-4o with chain-of-thought reasoning (Appendix \ref{app:doc-classification-prompt}), we categorized all API documentation in the API Extraction Benchmark. Out of the total, 24 documents were classified as Fully Organized, 134 as Semi-Organized, and 9 as Unorganized. Figure \ref{app:fig:api-webpage-example} provides examples of API documentation in each category. Notably, none of the documentation is fully standardized, and Semi-Organized and Unorganized documents frequently lack clarity or essential API details.

\section{JSON-To-Tool generation}
\label{app:tool-generator}
JSON-To-Tool generation is mostly about engineering. As the documentations are written in various formats, multiple conditions need to be considered, especially when handling URLs. It is common for the API server to only support a specific input pattern, which is not mentioned in the documentation or the error message. To make the autogenerated tool more robust, we considered the following procedures in our tool generator:

\textbf{parameter handling} Documentations use various ways to represent path parameters. We set matching rules to find commonly used patterns such as "\verb|:param|", "\verb|{param}|", "\verb|<param>|" etc.

\textbf{encoding correction} To pass some special characters such as "+" or "=", the URL will use an encoding method known as percent encoding. Usually this won't be an issue, but for some APIs this need to be done before passing the parameters.

\textbf{required parameter checking} As APIs may not necessarily return the error information, we added a required parameter validation step in the generated tool so that we can report any missing parameter errors to the AI agent.

\section{Tool Error Types and Causes} \label{app:error-validation}
Errors are categorized below:

 \textbf{Incomplete URL:} Occasionally, the URL of a tool is incomplete, resulting in failed requests. These errors can be classified into two subtypes: \textbf{Missing Endpoint Path} or \textbf{Missing Base URL}. By examining the corresponding API documentations, we found that endpoint paths were usually provided, but base URLs were often missing.

 \textbf{Request Errors:} Request errors are complex and challenging to diagnose, as status codes alone do not clearly indicate whether the issue originates from the server or client side. A status code of 200 ("OK") guarantees valid communication between the client and server. To further validate the response content in such cases, we use an GPT-4o based evaluator (Appendix \ref{app:validation-prompt}). If the evaluator returns "pass", the tool is labeled as \textbf{Passed Validation}; otherwise, it is labeled as \textbf{Failed Validation}. For other status codes or unexpected cases, the tool is classified under \textbf{Abnormal Response}.

\textbf{Incorrect Parameter Values:} If a tool throws an exception that does not fall into any of the previously mentioned scenarios, it indicates that the tool was very likely called with invalid parameters. Specifically:
    \begin{itemize}
        \item If a required parameter is missing, the error is classified as \textbf{No Parameter Value}.
        \item If all required parameters are provided but the tool still fails, the issue likely stems from an incorrect example value, and the error is classified as \textbf{Wrong Parameter Value}.
    \end{itemize}

For all error types other than \textbf{Passed Validation}, we group them into four main categories: \textbf{C1} Missing API Documentation Details, \textbf{C2} Incorrectly Extracted URL Path, \textbf{C3} Incorrect Parameter Values, and \textbf{C4} Server-Side Errors. For each category, we provide a range of possible error diagnosis, from the most conservative to the most aggressive (see Appendix \ref{app:tab:error-estimation}). 

\begin{table*}[htbp]
\resizebox{\textwidth}{!}{
\begin{tabular}{lll}
\toprule
\multicolumn{1}{c}{\textbf{Category}}             & \multicolumn{1}{c}{\textbf{Conservative Estimate}}                                                 & \multicolumn{1}{c}{\textbf{Aggressive Estimate(Additional Terms Only)}}                          \\
\cmidrule(rl){1-1} \cmidrule(rl){2-2} \cmidrule(rl){3-3}
Missing API Documentation Details & 0                                                                                         & Missing Base URL+No Parameter Value   \\
Incorrectly Extracted URL Path    & Missing Endpoint Path                                           & Missing Base URL                                              \\
Incorrect Parameter Values   & Wrong Parameter Value+Failed Validation & No Parameter Value+Abnormal Response  \\
Server-Side Error                              & 0                                                                                         & Failed Validation+Abnormal Response \\
\bottomrule
\end{tabular}
}
\caption{Estimation of error causes}
\label{app:tab:error-estimation}
\end{table*}

We aim to automatically fix the tools that either failed in the validation or can't be validated due to missing parameter examples. We find that most of the errors are caused by \textbf{C3}. We investigate the failed tools and find the example parameters from such tools are either missing or the parameter is not filled properly due to low-quality documentation. This inspires us to develop an approach to produce high-quality parameter values.

\section{Glycoscience Agent}
\subsection{Contribution to Glyco Research Community}

Our AI agent provides natural language interfaces for researchers to access web services (e.g., databases and utility APIs) without requiring technical expertise, which greatly facilitates the usage of online resources and benefits researchers in several ways:

\textbf{(1) Automatic cross-database integration\quad} No single database fulfills all information needs due to their specific focuses. For example, GlyTouCan catalogs glycan structures, KEGG GLYCAN maps pathways and reactions, and PubChem offers general molecular data. Our AI agent integrates these diverse sources for seamless information access. 

\textbf{(2) Tool synergy\quad} Individual tools are often limited to specific scenarios, but our AI agent enhances their applicability by integrating them into cohesive workflows, including ID conversion, database querying, data normalization, visualization, and resolution of inconsistencies. Appendix Table \ref{app:tab:glycanid} shows an example of a glycan being represented using various formats. This integration broadens the applicability of certain tools. For instance, the GLYCAM 3D visualization tool, initially limited to GLYCAM strings, now supports multiple glycan formats through automated conversion.

\textbf{(3) Facilitate the development and adoption of new services\quad} Using Doc2Agent, our AI agent can easily adopt new APIs and datasets, keeping its tools up-to-date. This enables researchers to prioritize scientific exploration over technical integration. In addition, the AI agent can serve as a platform for developing, sharing, and publishing applications, fostering dissemination and collaboration.

\subsection{Parameters in Glycoscience APIs} \label{app:glyco-api-complexity}
\begin{table}[htbp]
\centering

\renewcommand{\arraystretch}{1.5}

\begin{tabular}{ccc}
\toprule
\multicolumn{3}{c}{\textbf{String Representation}}  \\ \cmidrule(rl){1-3}
\multicolumn{1}{c}{\textbf{IUPAC Condensed}} & \multicolumn{2}{c}{Fuc(a1-2)Gal(b1-3){[}Fuc(a1-4){]}GlcNAc(b1-}     \\
\multicolumn{1}{c}{\textbf{GLYCAM}}          & \multicolumn{2}{c}{LFucpa1-2DGalpb1-3{[}LFucpa1-4{]}DGlcpNAcb1-OH}   \\ \bottomrule \toprule
\multicolumn{2}{c}{\textbf{Database ID}} & \multicolumn{1}{c}{\textbf{Glycan Name}} \\ \cmidrule(rl){1-2} \cmidrule(rl){3-3}
\multicolumn{1}{c}{\textbf{GlyToucan ID}} & \multicolumn{1}{c}{G00048MO} & \multirow{2}{*}{Lewis b} \\
\multicolumn{1}{c}{\textbf{PubChem ID}} & \multicolumn{1}{c}{45480569} & \multicolumn{1}{c}{} \\
\bottomrule
\end{tabular}
\caption{Example of different representations of Glycan \textit{Lewis b}. The table shows various ways to represent the glycan, demonstrating the complexity of glycan reference.}
\label{app:tab:glycanid}
\end{table}

\begin{table}[htbp]
\centering

\begin{tabular}{ccc}
\toprule
\textbf{Database} & \textbf{Reference ID} & \textbf{Primary Usage}                                                            \\ 
\midrule
GlyTouCan         & GlyTouCan ID          & \begin{tabular}[c]{@{}c@{}}Glycan Structure\\ GlyTouCan ID for API Calling\end{tabular} \\ \midrule
KEGG GLYCAN       & KEGG ID               & \begin{tabular}[c]{@{}c@{}}KEGG Pathyway\\ Reaction\end{tabular}                        \\ \midrule
GlyGen            & GlyTouCan ID          & \begin{tabular}[c]{@{}c@{}}Publication\\ Cross Reference\end{tabular}                   \\ \midrule
O-GlcNAc          & UniProtKB ID          & Protein O-GlcNAcylation Data                                                            \\ \midrule
PubChem           & PubChem CID           & Chemical Properties                                                                     \\ \midrule
Unilectin         & Unilectin ID          & Get Lectin and Ligand Information                                                       \\ 
\bottomrule
\end{tabular}

\caption{Glycan databases covered in glyco agent}
\label{app:tab:glycodataset}
\end{table}

Table \ref{app:tab:glycanid} and \ref{app:tab:glycodataset} shows the complexity of database entries and representations in glyco databasee, demonstrating the importance of infering correct parameter values for domain-specified agent.

To simulate missing parameter information, we applied a leave-one-API-out setup: for each API, we masked all corresponding parameter values and related entries in the database. Our parameter inference algorithm then generated candidate values using the remaining data. We evaluated the success of each inferred value using an LLM-as-a-judge framework, assessing whether the resulting tool call executed correctly.

\begin{table}[htbp]
\centering
\begin{tabular}{ccc}
\toprule
\textbf{} & \textbf{Ours} & \textbf{GPT-4o} \\ \midrule
Pass@10   & 33            & 17    \\ \bottomrule         
\end{tabular}

\caption{Successful Parameter Value Guess in glyco-material APIs}
\label{tab:experiment-extract-compare}
\end{table}

\label{app:parameter-value-infernence-comparison}
\noindent \textbf{Results} As shown in Table \ref{tab:experiment-extract-compare}, by using the leave-one-API-out test setting, our approach successfully infers parameter values for 33 tools. Meanwhile, GPT-4o managed to find the parameter values for only 17 tools. The results clearly indicate the superiority of our approach over GPT-4o.

\subsection{Deployment of the Research Agent}
\begin{figure}[htbp]
\centering
\includegraphics[width=0.8\linewidth]{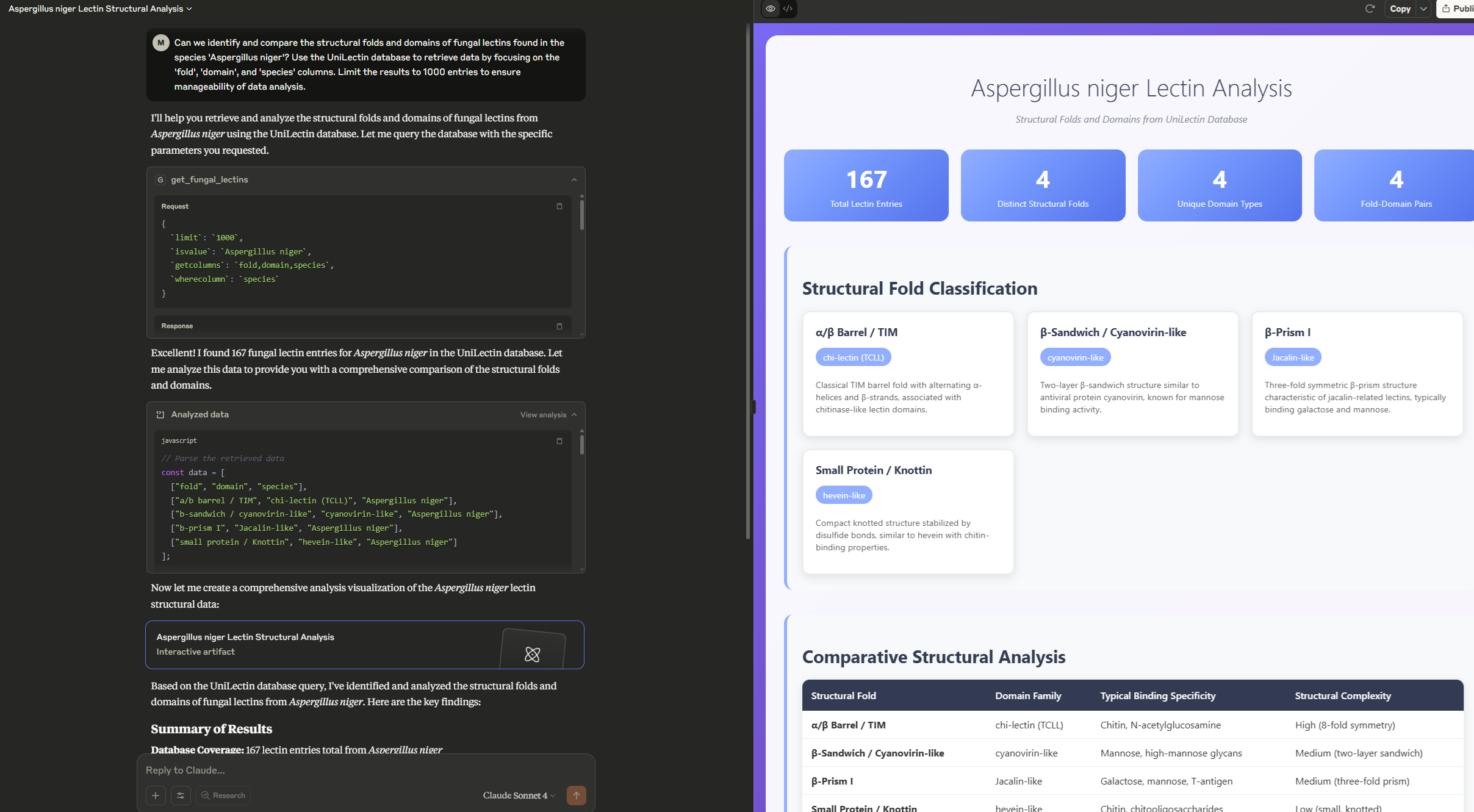}
\caption{Example of a research agent produced by Doc2Agent aggregated into Claude Desktop. It can solve domain-specific questions with tools and generate reports in various formats.}
\label{app:fig:glyco-agent-example}
\end{figure}

A research agent can be easily deployed through any agentic framework. Figure~\ref{app:fig:glyco-agent-example} shows an example of the aggregation of the MCP server produced by Doc2Agent framework and Claude Desktop UI.

\section{WebArena Evaluation for API-agent} \label{app:evluation}
We modify the vanilla WebArena evaluator to tackle inaccurate results for better evaluation for API using agents. 

\subsection{Intend Confirmation}
A common source of false positives occurs when the agent fails to execute the intended task but inadvertently includes an output that satisfies the string match criteria. To ensure the agent’s behavior aligns with the task intent, we employ a two-step string match evaluation process:
(1) The agent’s output is first evaluated using the standard WebArena string match evaluator to confirm the presence of the target reference string.
(2) An LLM then reviews the agent’s reasoning path to determine whether the task was genuinely completed or if the match was incidental.

An example of such a false positive is shown below:

\begin{tcolorbox}[title=WebArena False Positive Case]
WebArena Task 22:
\begin{lstlisting}[style=colorEX, language=PyBrIM]
'''List out reviewers, if exist, who mention about under water photo'''
\end{lstlisting}
The vanilla evaluator validate the answer with:
\begin{lstlisting}[style=colorEX, language=PyBrIM]
"fuzzy_match": "N/A" ("There is no review about under water photo")
\end{lstlisting}
In the agent log, the agent ends up getting struck and accidentally output nothing, which allows it to pass the test:
\begin{lstlisting}[style=colorEX, language=PyBrIM, basicstyle=\ttfamily\tiny]
Step 4:
To list out reviewers who mention "under water photo" for the product with SKU B001D0G57S, I will follow these steps:

1. **Check for an API to retrieve product reviews**: I will look for an API that can fetch reviews for a specific product.
2. **Fetch the reviews using the API**: If an API is available, I will use it to get the reviews for the product with SKU B001D0G57S.
3. **Filter reviews mentioning "under water photo"**: I will filter the reviews to find those that mention "under water photo".
4. **Verify the results using web browsing**: I will navigate to the product page and manually verify the reviews to ensure the results are correct.

First, let's check for an API that can retrieve product reviews.
...
Step 12:
Too many errors encountered. Task failed.
Step 13:
Too many errors encountered. Task failed.
Step 14:
Too many errors encountered. Task failed.
\end{lstlisting}
Vanilla WebArena Evaluator
\begin{lstlisting}[style=colorEX, language=PyBrIM]
correct: true
\end{lstlisting}
API-specified Evaluator
\begin{lstlisting}[style=colorEX, language=PyBrIM]
correct: false
\end{lstlisting}
\end{tcolorbox}

\begin{tcolorbox}[title=String-match LLM-as-a-Judge Prompt]
We use an LLM to check the agent log.
\begin{lstlisting}[style=colorEX, language=PyBrIM]
'''You are evaluating an agent's performance on the task: 
{task_intent}

The agent's success is measured by these requirements: 
{evaluator_str}

Below are the execution logs. Determine whether the agent has successfully completed the task based on the requirements.
{extracted_log}

IMPORTANT: Your evaluation should be based on whether the agent achieved the task's objective. Focus on the logical reasoning path and the final outcome.

AGENT'S FINAL ANSWER: 
{final_answer}

Provide your assessment in this exact format:
REASON: [One sentence explaining your decision]
DECISION: [success/fail]'''
\end{lstlisting}
\end{tcolorbox}

\subsection{Adaption for API-agent}
The \verb|url_match| and \verb|program_html| evaluator in WebArena assesses agent actions by examining URLs or changes in webpage elements—such as verifying whether a string was correctly entered into a search field. However, this type of interaction can be bypassed entirely by API-based agents. For example, rather than add search strings into the search bar and clicking a \verb|search| button, an API agent may directly invoke a search endpoint which uses a different URL to the button link. Such behavior is not detectable by the original evaluator.

To address this limitation, we extend the evaluator to support API-calling functions by enabling it to match relevant information across API requests, responses, and webpage content. Based on this matched information, we incorporate an LLM-based evaluator to determine:
(1) whether the API agent’s action trajectory is functionally equivalent to the intended \verb|url_match| or \verb|program_html| behavior, and
(2) whether the agent successfully completes the task. An example is provided below:

\begin{tcolorbox}[title=WebArena API use example]
WebArena Task 395:
\begin{lstlisting}[style=colorEX, language=PyBrIM, basicstyle=\ttfamily\tiny]
'''Fork the Pytorch GAN repo with most stars.'''
\end{lstlisting}
The vanilla evaluator validate the answer with:
\begin{lstlisting}[style=colorEX, language=PyBrIM, basicstyle=\ttfamily\tiny]
"__GITLAB__/byteblaze/PyTorch-GAN": must_include("Pytorch-GAN")
\end{lstlisting}
API-agent action
\begin{lstlisting}[style=colorEX, language=PyBrIM, basicstyle=\ttfamily\tiny]
'''Step 5:
To fork the Pytorch GAN repository with the most stars on GitLab, we need to follow these steps:
1. List the available tools in the 'projects' category to find tools that can help us search for projects.
2. Use the appropriate tool to search for Pytorch GAN repositories and identify the one with the most stars.
3. Use the project ID of the identified repository to fork it.
Let's start by listing the tools in the 'projects' category to find the right tool for searching projects.
<observation not shown>
Step 6:
We have a list of tools available in the 'projects' category. To find the Pytorch GAN repository with the most stars, we should use the `list_projects` tool to search for relevant projects and then identify the one with the most stars.
Let's first get the documentation for the `list_projects` tool to understand how to use it effectively.
<observation not shown>
Step 7:
The `list_projects` tool allows us to search for GitLab projects with various filtering options. We can use the `search` parameter to look for Pytorch GAN projects and the `order_by` and `sort` parameters to sort the results by the number of stars in descending order.
Let's proceed to call the `list_projects` function to search for Pytorch GAN projects and identify the one with the most stars.
<observation not shown>
Step 8:
It seems that the `order_by` parameter does not accept `star_count` as a valid value. Let's check the documentation again to see what valid values are available for the `order_by` parameter. We might need to use a different approach to identify the project with the most stars.
I'll review the documentation for the `list_projects` tool again to find the correct `order_by` options.
<observation not shown>
Step 9:
The documentation indicates that the `order_by` parameter defaults to `created_at`, but it doesn't specify other valid values. Since `star_count` is not valid, we should try using the default `created_at` and then manually identify the project with the most stars from the results.
Let's call the `list_projects` function again without specifying `order_by` and `sort` to retrieve the projects and then find the one with the most stars.
<observation not shown>
Step 10:
The response was truncated, so we need to retrieve the full response to analyze all the projects and identify the one with the most stars. Let's do that now.
<observation not shown>
Step 11:
We have retrieved the full response, and it contains details about the Pytorch GAN projects. Now, let's identify the project with the most stars and proceed to fork it.
The project with the most stars is "PyTorch-GAN" by Erik Linder-Noren, with 45 stars. We will use the `fork_project` tool to fork this project.
Let's proceed to get the documentation for the `fork_project` tool to understand how to use it.
<observation not shown>
Step 12:
The `fork_project` tool allows us to create a fork of an existing project. We need to provide the `project_id` of the project we want to fork. We can also specify additional parameters like `namespace_path`, `name`, and `description` if needed.
Let's proceed to fork the "PyTorch-GAN" project with ID 86.
<observation not shown>
Step 13:
The "PyTorch-GAN" project has been successfully forked. The forked project is now under the namespace "Byte Blaze" with the project ID 195.\n\nFinish[The Pytorch GAN repo with the most stars has been successfully forked.]
\end{lstlisting}
LLM Reason
\begin{lstlisting}[style=colorEX, language=PyBrIM, basicstyle=\ttfamily\tiny]
The agent successfully forked the \"PyTorch-GAN\" project, as indicated by the status code 201 and the confirmation that the forked project is under the namespace \"Byte Blaze\" with a new project ID.'''
\end{lstlisting}
LLM decision
\begin{lstlisting}[style=colorEX, language=PyBrIM, basicstyle=\ttfamily\tiny]
correct: true
\end{lstlisting}
\end{tcolorbox}

\begin{tcolorbox}[title=Program html LLM-as-a-Judge Prompt]
We use an LLM to check the agent log.
\begin{lstlisting}[style=colorEX, language=PyBrIM, basicstyle=\ttfamily\tiny]
'''You are evaluating an API agent's performance on the task: {task_intent}

The reference implementation accesses {ref_urls_str}, but this agent uses a different API ecosystem. Your goal is to determine if the agent's API calls accomplish the same functional objective, regardless of specific endpoints used.

**Evaluation Criteria:**
1. **Functional Alignment**: Do the API calls serve the same purpose as the reference URL?
2. **Parameter Relevance**: Are the parameters extracted from the query and used appropriately?
3. **Logical Sequence**: Do the API calls follow a reasonable order to accomplish the task?
4. **Error Handling**: Does the agent handle failures gracefully and attempt alternatives?

**Assessment Guidelines:**
- SUCCESS if the agent's API calls would reasonably accomplish the same goal as the reference URL
- SUCCESS if the agent uses equivalent but different endpoints (e.g., different weather APIs for weather queries)  
- SUCCESS if the agent makes multiple related calls that collectively achieve the objective
- FAIL only if the API calls are clearly unrelated to the task or would not achieve the intended outcome
- Consider the agent's reasoning process from the logs, not just the final API calls

**Parameter Analysis:**
{overlap_functions}
Reference parameters: {ref_parameter}
Agent parameters: {agent_parameter}

**Agent Execution:**

Execution log:
{extracted_log}

**Your Assessment:**
REASON: [One sentence explaining whether the agent's approach would accomplish the same objective as the reference URL, considering functional equivalence rather than exact matching]
DECISION: [success/fail]'''
\end{lstlisting}
\end{tcolorbox}

\section{Direct Tool Generation Example}
\label{app:API-extraction-example}
\begin{tcolorbox}[title=Tool Generation Example]
We extract web contents into JSONs first. \href{https://pokemontcg.io/?ref=apilist.fun}{the API documentation(HTML) page}, while the output is a JSON string following our API-extraction Schema. Below is the input example:
\begin{lstlisting}[language={},basicstyle=\ttfamily\tiny]
# OSRM HTTP Router
## General options
All OSRM HTTP requests use a common structure.
The following syntax applies to all services, except as noted.
### Requests
```endpoint
GET http://http://ec2-3-129-135-45.us-east-2.compute.amazonaws.com:{profile}/{service}/v1/test/{coordinates}[.{format}]?option=value&option=value
```
| Parameter | Description |
| --- | --- |
| `profile` | Mode of transportation. One of the following three values: `5000` for car (driving), `5001` for bicycle (biking), and `5002` for foot (walking). |
| `service` | One of the following values: [`route`](#route-service), [`nearest`](#nearest-service), [`table`](#table-service), [`match`](#match-service), [`trip`](#trip-service), [`tile`](#tile-service) |
| `coordinates`| String of format `{longitude},{latitude};{longitude},{latitude}[;{longitude},{latitude} ...]` or `polyline({polyline}) or polyline6({polyline6})`. |
| `format`| `json` or `flatbuffers`. This parameter is optional and defaults to `json`. |
......
\end{lstlisting}

The output example shows as follows:

\begin{lstlisting}[style=colorEX, language=PyBrIM,basicstyle=\ttfamily\tiny]
{
    "title": "OSRM HTTP Router API Documentation",
    "endpoints": [
        {
            "name": "General Request",
            "description": "All OSRM HTTP requests use a common structure.",
            "method": "GET",
            "url": [
                "http://ec2-3-129-135-45.us-east-2.compute.amazonaws.com:{profile}/{service}/v1/test/{coordinates}[.{format}]?option=value&option=value"
            ],
            "headers": [],
            "required_parameters": [
                {
                    "name": "profile",
                    "type": "string",
                    "description": "Mode of transportation. One of the following three values: '5000' for car (driving), '5001' for bicycle (biking), and '5002' for foot (walking).",
                    "default": null,
                    "example": "5000"
                },
                {
                    "name": "service",
                    "type": "string",
                    "description": "One of the following values: 'route', 'nearest', 'table', 'match', 'trip', 'tile'.",
                    "default": null,
                    "example": "route"
                },
                ...
            ],
            "optional_parameters": [
                {
                    "name": "format",
                    "type": "string",
                    "description": "'json' or 'flatbuffers'. This parameter is optional and defaults to 'json'.",
                    "default": "json",
                    "example": "json"
                }
            ]
        },
}
\end{lstlisting}
\end{tcolorbox}

\begin{tcolorbox}[title=Tool Generation Example]
By appling the JSON-To-Tool generation script, we can convert the structured API information into an excutable tool. In this work, a tool is a python function which can be called by the LLM. A typical tool example for 'Pokemon TCG API' is shown below. \label{app:tool-example}
\begin{lstlisting}[style=colorEX, language=PyBrIM]
import requests
import json
from urllib.parse import quote

def general_request(profile=None, service=None, coordinates=None, format='json', **kwargs):
    assert profile is not None, 'Missing required parameter: profile'
    assert service is not None, 'Missing required parameter: service'
    assert coordinates is not None, 'Missing required parameter: coordinates'
    
    base_url = f"http://ec2-3-129-135-45.us-east-2.compute.amazonaws.com:{profile}/{service}/v1/test/{coordinates}"
    
    
    response = requests.get(url=base_url, params=kwargs, timeout=50)
    return response

if __name__ == '__main__':
    r = general_request(profile='5000', service='route', coordinates='13.388860,52.517037;13.397634,52.529407', format='json')
    r_json = None
    try:
        r_json = r.json()
    except:
        pass
    result_dict = dict()
    result_dict['status_code'] = r.status_code
    result_dict['text'] = r.text
    result_dict['json'] = r_json
    result_dict['content'] = r.content.decode("utf-8")
    print(json.dumps(result_dict, indent=4))
\end{lstlisting}
\end{tcolorbox}

\begin{tcolorbox}[title=Refined Tool Example]
After validation and refinement, bugs will be fixed and \verb|__doc__| will be added to the generated python function based on the API response.
\begin{lstlisting}[style=colorEX, language=PyBrIM, basicstyle=\ttfamily\tiny]
import requests
import json
from urllib.parse import quote

def general_request(profile=None, service=None, coordinates=None, format='json', **kwargs):
    """
    Make a request to the OSRM HTTP Router API.
    
    Parameters:
    -----------
    profile : str
        Mode of transportation. One of the following values:
        - '5000' for car (driving)
        - '5001' for bicycle (biking)
        - '5002' for foot (walking)
    
    service : str
        One of the following values:
        - 'route': Find the fastest route between coordinates
        - 'nearest': Snap a coordinate to the street network
        - 'table': Compute the duration matrix between coordinates
        - 'match': Match GPS points to the road network
        - 'trip': Solve the Traveling Salesman Problem
        - 'tile': Generate Mapbox Vector Tiles
    
    coordinates : str
        String of format '{longitude},{latitude};{longitude},{latitude}[;{longitude},{latitude} ...]'
        or 'polyline({polyline})' or 'polyline6({polyline6})'.
        
    format : str, optional (default='json')
        Response format, either 'json' or 'flatbuffers'.
    
    **kwargs : dict
        Additional query parameters to pass to the API.
        Common options include:
        - overview: 'simplified', 'full', or 'false'
        - steps: 'true' or 'false'
        - annotations: 'true', 'false', or specific values
        - geometries: 'polyline', 'polyline6', or 'geojson'
    
    Returns:
    --------
    requests.Response
        The response from the OSRM API.
    
    Examples:
    ---------
    >>> # Get a route between two points
    >>> response = general_request(
    ...     profile='5000',
    ...     service='route',
    ...     coordinates='13.388860,52.517037;13.397634,52.529407',
    ...     overview='false'
    ... )
    
    >>> # Get a table of durations between three points
    >>> response = general_request(
    ...     profile='5000',
    ...     service='table',
    ...     coordinates='13.388860,52.517037;13.397634,52.529407;13.428555,52.523219',
    ...     annotations='duration'
    ... )
    """
    assert profile is not None, 'Missing required parameter: profile'
    assert service is not None, 'Missing required parameter: service'
    assert coordinates is not None, 'Missing required parameter: coordinates'
    
    base_url = f"http://ec2-3-129-135-45.us-east-2.compute.amazonaws.com:{profile}/{service}/v1/test/{coordinates}"
    
    if format and format != 'json':
        base_url += f".{format}"
    
    response = requests.get(url=base_url, params=kwargs, timeout=50, verify=False)
    return response

if __name__ == '__main__':
    r = general_request(profile='5000', service='route', coordinates='13.388860,52.517037;13.397634,52.529407', format='json')
    r_json = None
    try:
        r_json = r.json()
    except:
        pass
    result_dict = dict()
    result_dict['status_code'] = r.status_code
    result_dict['text'] = r.text
    result_dict['json'] = r_json
    result_dict['content'] = r.content.decode("utf-8")
    print(json.dumps(result_dict, indent=4))
\end{lstlisting}
\end{tcolorbox}

\section{API-extraction Schema} \label{app:json-schema}
\begin{tcolorbox}[title=API-extraction Schema]
We defined the Pydantic schema to extract API information.
\begin{lstlisting}[style=colorEX, language=PyBrIM]
class Parameters(BaseModel):
    name: str = Field(description="Name of the parameter")
    type: Optional[str] = Field(description="Type of the parameter")
    description: Optional[str] = Field(description="Description of the parameter. If the parameter is categorical, please list all possible values.")
    default: Optional[Any] = Field(
        None,
        description="Default value of the parameter")
    example: Optional[Any] = Field(
        description="Example value of the parameter")


class Endpoint(BaseModel):
    name: str = Field(description="Name of the endpoint")
    description: Optional[str] = Field(
        description="Description of the endpoint")
    method: str = Field(description="Method of the endpoint")
    url: Union[str, List[str]] = Field(description="URL of the endpoint, start with http:// or https://")
    headers: Optional[List] = Field(
        default=[], description="Headers of the endpoint")
    required_parameters: Optional[List[Parameters]]
    optional_parameters: Optional[List[Parameters]]


class Api_json(BaseModel):
    title: Optional[str] = Field(description="Title of the API")
    endpoints: List[Endpoint]
\end{lstlisting}
\end{tcolorbox}

\section{LLM Prompts}
\subsection{Doc2Agent Prompt} \label{app:validation-prompt}
\begin{tcolorbox}[title=API Response Validation Prompt]
The prompt for the model is defined as
\begin{lstlisting}[style=colorEX, language=PyBrIM]
"""
Given an API description, response, and Python code, classify the response type:

- information: Valid response containing useful data as expected
- code_error: Error due to bugs in the Python code (syntax, logic, url path doesn't match, etc.)
- server_error: Server-side error (500, service unavailable, authentication error, etc.)
- request_error: Invalid operation due to bad parameters or unsupported operation (only choose this when other errors are not applicable and you think the code is correct but the response is an error)

API Description: {description}
API Response: {response}
Code: {code}
"""
\end{lstlisting}
The expected output is:
\begin{lstlisting}[style=colorEX, language=PyBrIM]
class Classification(BaseModel):
    response_type: str = Field(..., enum=['information', 'code_error', 'server_error', 'request_error'])
\end{lstlisting}
\end{tcolorbox}

\begin{tcolorbox}[title=API Refinement Prompt]
To refine a tool, we use the following template
\begin{lstlisting}[style=colorEX, language=PyBrIM, basicstyle=\ttfamily\tiny]
"""
You are a Python debugging expert specializing in code analysis and correction. Your will be provided with:
1. Error information.
2. API Json documentation: The API documentation includes all endpoints information. Please refer to the most relevent one based on the similarity of the Python function name and the endpoint's name.
3. Python code containing the function that interacts with APIs: The code may contain syntax errors, parameter issues, or other problems that prevent it from running correctly.
4. Parameters: The parameter examples that are passed to the function.
5. Configuration: The API configuration including base URL, headers, authentication details, and testing information.

Your task is to identify and fix errors in the provided code based on the error information and API documentation.

Requirements:
- IMPORTANT: Return ONLY the corrected code without explanations. The code you answered should be run successfully in a Python environment without any other human modifications.
- DON'T return the code with the wrap of ```python```.
- Ensure the code includes import statements for any necessary libraries.
- If the parameter examples are empty or not given, you should guess the correct values based on the API documentation and include them.
- If the configuration includes an "info" section with user_name, user_id, project_id, etc., use these values as realistic test parameters when the function needs user-specific data. These are real test values that should work with the API.
- Please write the doc after refineing the code so that I can directly call code.__doc__. Be sure to include examples of the parameters in the docstring.
- Please check the parameters and make sure they are passed to the url correctly. Add optional parameters to allow custumizerble use of the tool when necessary.
- Use the provided configuration for base URL, headers, and authentication. Replace any hardcoded URLs or headers with the configuration values.
- You should **NEVER** edit the result_dict in main function.

Error Information:
{error}

API Documentation:
{api_doc}

Configuration:
{config}

Code to fix:
{code}

Candidate Parameter Values:
{params}
"""
\end{lstlisting}
\end{tcolorbox}

\begin{tcolorbox}[title=Parameter Value Inference Prompt]
The prompt for direct parameter value inferencing is
\begin{lstlisting}[style=colorEX, language=PyBrIM]
'''
You will be provided with the information of an API and its parameters. The example values of the parameters are missing. You need to guess the parameter values.
You may have failed severl times before. If you guess with similar values, you may fail again. Please be innovative and try different values and formats.

Your previous failed guesses:
***history start
{history}
***history end

API Description:
{description}

Parameter Description:
{param_description}

Your Guess:
'''
\end{lstlisting}
\end{tcolorbox}

\subsection{API Documentation Classification Prompt} \label{app:doc-classification-prompt}
\begin{tcolorbox}[title=API Response Validation Prompt]
The prompt for the model is
\begin{lstlisting}[style=colorEX, language=PyBrIM]
'''
You need to group the API documentation with the following standards:

Fully Organized: The documentation follows a well defined template, most likely to be from an API platform. It is well-structured, clear, and easy to understand. It includes detailed descriptions, example code, and explanations of how to use the API.
Semi-Organized: Lacks some structure, but still includes most of the necessary information. It may be missing some examples or descriptions, making it slightly more difficult to understand how to use the API.
Unorganized: Missing example or description, or the structure is unclear, making it difficult to understand how to use the API.

===
API Documentation:
{API_DOC}
'''
\end{lstlisting}
The output of the mode is defined as
\begin{lstlisting}[style=colorEX, language=PyBrIM]
# Output Class Structure
class Classification(BaseModel):
    analysis: str = Field(..., description="The analysis of the API documentation. Make it within 300 characters.")
    category: str = Field(..., enum=["Fully Organized", "Semi-Organized", "Unorganized"])
\end{lstlisting}
\end{tcolorbox}

\subsection{LLM-as-a-Judge Prompt} \label{app:llm-as-a-judge-prompt}
\begin{tcolorbox}[title=Recheck WebArena Tasks]
We use LLM to double check the WebArena Tasks to evaluate if the agent actually complete the task.
\begin{lstlisting}[style=colorEX, language=PyBrIM]
'''
You are an expert evaluator. 
Given a task log and the reference answer.
Your job is to compare the log and the reference answer and then determine if the log shows that its job is done. 
Answer, respond with ONLY one word: 'true' or 'false'.

Note: 
Focus on the content of the log and the reference answer. 
Check if the result has the same meaning as the reference answer.
For shopping tasks, check if the correct cart operations were performed or the correct product information was retrieved.

### Reference answer
{}

### Task log
{log_text}

### Your decision
'''
\end{lstlisting}
\end{tcolorbox}

\begin{tcolorbox}[title=Evaluator for Glycan dataset]
\begin{lstlisting}[style=colorEX, language=PyBrIM, basicstyle=\ttfamily\tiny]
'''
"""
You are an expert glycomics researcher evaluating an AI agent's performance on a glycan research task.

**RESEARCH QUESTION:**
{question}
**EXPECTED API TRAJECTORY:**
{expected_trajectory}
**EXPECTED ANSWER:**
{expected_answer}
**AGENT'S COMPLETE INTERACTION LOG:**
{agent_log}
**AGENT'S FINAL RESPONSE:**
{agent_final_response}

**EVALUATION CRITERIA:**
Evaluate whether the agent successfully completed the research task by considering ALL of the following aspects holistically:

1. **API Usage Appropriateness**: Did the agent use the right sequence of glycan APIs to address the research question? The agent should have used APIs that align with the expected trajectory, though exact matching is not required if the alternative approach is scientifically valid.

2. **Scientific Accuracy**: Are the obtained results scientifically correct and meaningful for glycomics research? Consider:
   - Correct interpretation of glycan structures and formats (WURCS, IUPAC, etc.)
   - Proper understanding of protein-glycan interactions
   - Accurate use of database cross-references
   - Valid biochemical reasoning

3. **Completeness**: Did the agent obtain all the information requested in the question? Check if:
   - All parts of multi-part questions were addressed
   - Required data fields were retrieved
   - Cross-references were properly resolved

4. **Data Integration**: If multiple API calls were needed, did the agent properly combine and interpret the results? Look for:
   - Logical workflow progression
   - Proper data passing between API calls
   - Meaningful synthesis of results from different sources

5. **Research Workflow**: Did the agent follow a logical research workflow consistent with glycomics best practices?
   - Started with appropriate tool discovery
   - Used documentation when needed
   - Handled errors gracefully
   - Drew appropriate conclusions

**IMPORTANT GUIDELINES:**
- Focus on whether the agent achieved the research objective, not just exact trajectory matching
- Consider that there may be multiple valid approaches to solve the same research question
- Evaluate the scientific validity of the final answer in the context of glycomics
- Account for API errors or limitations that may have affected the agent's performance
- Consider partial credit for agents that made significant progress but didn't fully complete the task
- Be especially attentive to:
  * Correct format conversions (e.g., IUPAC to WURCS)
  * Proper interpretation of molecular structures
  * Valid cross-database references
  * Meaningful biological insights

**RESPONSE FORMAT:**
Provide your evaluation in this exact format:

DECISION: [SUCCESS/FAILURE]

REASONING: [Detailed explanation of your decision, covering the key evaluation criteria and specific observations from the agent's performance. Include specific examples from the agent's log that support your decision.]

CRITICAL_ISSUES: [List any major problems that led to failure, or "None" if successful]

SCIENTIFIC_ACCURACY: [Assessment of the scientific validity of the results, including any concerns about data interpretation or biochemical reasoning]

API_USAGE_ASSESSMENT: [Evaluation of whether the agent used appropriate tools and followed a logical workflow]
'''
\end{lstlisting}
\end{tcolorbox}

\end{document}